%% file: main_arxiv_11_01.tex
\documentclass{article}
\usepackage{iclr2027_conference,times}
\usepackage{amsmath,amssymb,amsthm,mathtools}
\usepackage{booktabs,tabularx,array}
\usepackage{graphicx,xcolor,colortbl}
\usepackage{tikz}
\usepackage{float}
\usepackage[noend]{algorithmic}
\usepackage{wrapfig}
\usepackage{flafter}
\usepackage{hyperref,url}
\input{macros}

\title{HybridSpec: Throughput-Aware Speculative Decoding for Hybrid GDN--MoE Models}
\author{
  Haiyu Wang\textsuperscript{1,2}, 
  Sai Qian Zhang\textsuperscript{2}, 
  Zechun Liu\textsuperscript{1}, 
  Steven Li\textsuperscript{1} \\
  \textsuperscript{1}Meta AI \\
  \textsuperscript{2}New York University
}
\iclrfinalcopy
\hypersetup{hidelinks,pdftitle={HybridSpec: Throughput-Aware Speculative Decoding for Hybrid GDN--MoE Models},pdfauthor={
  Haiyu Wang\textsuperscript{1,2}, 
  Sai Qian Zhang\textsuperscript{2}, 
  Zechun Liu\textsuperscript{1}, 
  Steven Li\textsuperscript{1} \\
  \textsuperscript{1}Meta AI \\
  \textsuperscript{2}New York University
}}

\begin{document}
\maketitle

\thispagestyle{plain}
\pagestyle{plain}

\begin{abstract}
In speculative decoding, parallel drafters such as DFlash make drafting cheap, so target verification dominates each round. A central choice is the verification length, the number of tokens that the target processes in each round. Verifying more draft tokens can accept more of them, but work after the first rejection is wasted. For hybrid target models that combine Gated DeltaNet (GDN) and mixture-of-experts (MoE) layers, verification time grows much faster with this length than for dense models. In a single-request profile of Qwen3.6-35B-A3B with DFlash, it grew by 71\% from length 4 to 16, almost entirely in GDN and MoE layers, compared with 4\%--6\% for a dense Qwen3-4B target. Each GDN layer must update a recurrent state at every token and recover it at the end of the accepted prefix. The MoE cost depends on the experts that the tokens activate, which are known only once verification runs. We present \method{}, which selects the verification length on the GPU from draft-side acceptance estimates and measured target latency. Verification then runs this length plus one extra token up to the first MoE router, before any expert runs. There, a lightweight routing correction uses the observed expert assignments to decide whether to keep that token. The rest of verification runs only on the selected tokens. Block-parallel GDN kernels avoid storing a state after every token and reconstruct the accepted state of every GDN layer in one launch. In isolated benchmarks at verification lengths 4 to 16 with every draft token accepted, these kernels, including accepted-state reconstruction, are $1.54\times$--$4.04\times$ faster than the default kernels. With three GDN--MoE target models in single-request greedy decoding, \method{} is $1.75\times$--$1.88\times$ faster than autoregressive decoding and $1.30\times$--$1.33\times$ faster than DFlash.
\end{abstract}

\section{Introduction}
\label{sec:introduction}

\setlength{\columnsep}{8pt}%
\begin{wrapfigure}{r}{0.5\linewidth}
\vspace{-1.2em}
\centering
\includegraphics[width=\linewidth]{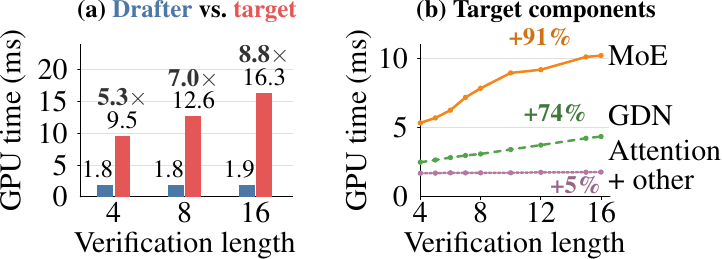}
\vspace{-1.8em}
\caption[Qwen3.6--DFlash GPU time across verification lengths]{\raggedright Profiled GPU time per verification call for \mbox{Qwen3.6-35B-A3B} with DFlash. (a)~Times in ms, with target/drafter ratios in bold. (b)~Bold percentages are increases from length 4 to 16.}
\label{fig:motivation-profile}
\vspace{-0.7em}
\end{wrapfigure}
Speculative decoding accelerates autoregressive generation by using a drafter to propose multiple tokens and a target model to verify them in a single forward pass, accepting draft tokens until the first rejection~\citep{leviathan2023fast,chen2023sampling}. Recent parallel and semi-autoregressive drafters, such as the diffusion-based DFlash, generate a block of draft tokens in one forward pass~\citep{hui2026peagle,chen2026dflash,cheng2026dspark}, making drafting inexpensive and shifting the main cost of each draft-and-verify round to verification, as Figure~\ref{fig:motivation-profile}a shows. This makes the verification length, i.e., the number of draft tokens processed by the target in each round, a key design choice. A longer verification length may accept more tokens, but working on any tokens after the first rejection incurs wasted computation.

Verification time grows much more rapidly with speculative length for the hybrid architectures used in recent open models such as Jamba, MiniMax-01, Kimi Linear, and Qwen3.5~\citep{lieber2024jamba,minimax2025minimax01,kimiteam2025kimilinear,qwen2026qwen35a3b} than for dense models. These architectures combine linear-attention or state-space layers, such as Gated DeltaNet (GDN)~\citep{yang2025gated}, with mixture-of-experts (MoE) layers. A GDN layer maintains a fixed-size recurrent state rather than a growing key--value cache, while an MoE layer routes each token to only a small subset of experts, 8 of 256 in the models we evaluate. In our profile of Qwen3.6-35B-A3B with a DFlash drafter at batch size one, increasing the verification length from 4 to 16 raises verification time by $71\%$, and Figure~\ref{fig:motivation-profile}b shows that almost all of the increase comes from the GDN and MoE layers. In contrast, verification time for a dense Qwen3-4B target increases by only $4\%$--$6\%$, since its batch-one cost is dominated by weight loading. For GDN, the default SGLang verifier~\citep{zheng2024sglang} updates the recurrent state at every token and stores a full copy after each update to enable rollback to the accepted prefix. For MoE, verification cost depends on expert reuse: additional tokens are inexpensive when they reuse already active experts but more costly when they activate new ones. Because experts are selected only when the target executes, a verification length chosen beforehand can only estimate this cost.


Existing length selectors, such as Sequoia~\citep{chen2024sequoia}, DSpark~\citep{cheng2026dspark}, \mbox{D-Cut}~\citep{liu2026dcut}, and EVICT~\citep{pan2026evict}, decide before verification from draft confidence or profiled costs. EcoSpec predicts draft-token experts with a separately trained model~\citep{xie2026ecospec}, and FASER prunes tokens at intermediate target layers~\citep{chen2026faser}. None of them uses the target's own expert assignments before the experts run.

\method{} selects the verification length based on measured target-model latency and acceptance estimates derived from the drafter's own signals. Since both GDN and MoE costs increase with the number of verified tokens, reducing the verification length lowers both costs. \method{} then verifies this length plus one additional draft token up to the first MoE router, which we call the router boundary. At this point, a lightweight routing correction uses the observed expert assignments to retain the extra token only when its expected acceptance benefit outweighs the additional verification latency. To further accelerate GDN, \method{} employs kernels that process all verification tokens in parallel without storing per-token recurrent states. Our contributions are as follows:
\begin{itemize}
 \setlength{\itemsep}{0pt}
 \setlength{\parsep}{0pt}
 \item \textbf{Throughput-aware length selection.} A GPU-resident policy chooses the verification length to maximize decode throughput, and a lightweight routing correction can add one token at the router boundary. In exact arithmetic, every selected length provably preserves the target's greedy outputs and sampling distribution.
 \item \textbf{Two-stage verification at the router boundary.} Verification pauses after the first MoE router and resumes only on the selected tokens, with no CPU round trip or recomputation.
 \item \textbf{Parallel GDN verification and state reconstruction.} GPU kernels verify all tokens of a round in parallel within each GDN layer and rebuild the accepted states of all GDN layers in one launch. In isolated benchmarks with every draft token accepted, they are \mbox{$1.54\times$--$4.04\times$} faster than the default SGLang kernels at lengths 4 to 16.
 \item \textbf{End-to-end speedups.} On three GDN--MoE target models under greedy decoding, \method{} is $1.75\times$--$1.88\times$ faster than autoregressive decoding and $1.30\times$--$1.33\times$ faster than vanilla DFlash, which verifies every draft token.
\end{itemize}

\section{Background and Related Work}
\label{sec:background}
\label{sec:related}

\subsection{Speculative decoding and adaptive verification}
\label{sec:adaptiverelated}

Speculative decoding verifies draft tokens with the target model, accepting them until the first mismatch, after which the target generates its own token~\citep{leviathan2023fast,chen2023sampling}. This process exactly preserves the target model's greedy output and sampling distribution. Tree-based methods verify multiple candidate continuations in a single target pass~\citep{miao2024specinfer}. Parallel drafters generate multiple tokens simultaneously using extra decoding heads~\citep{cai2024medusa}, a shared learnable hidden state~\citep{hui2026peagle}, or a lightweight diffusion model~\citep{chen2026dflash}. EAGLE-3 instead drafts autoregressively using features from multiple target layers~\citep{li2025eagle3}, while DSpark combines parallel and sequential drafting~\citep{cheng2026dspark}. In this work, we fix the parallel drafter and focus on reducing verification cost.

Adaptive methods dynamically determine how many draft tokens to verify or how speculation should be structured. SpecDec++ formulates draft-length selection as a stopping problem~\citep{huang2024specdecpp}, while online adaptation adjusts speculative configurations during inference~\citep{liu2025dropin}. EAGLE-2 constructs its draft tree based on draft confidence~\citep{li2024eagle2}, and Sequoia optimizes speculation trees using draft probabilities together with hardware costs~\citep{chen2024sequoia}. DSpark selects verification lengths from prefix-level acceptance estimates and an engine-specific throughput profile~\citep{cheng2026dspark}, while \mbox{D-Cut} jointly prunes draft tokens across a batch using draft confidence and a runtime cost model~\citep{liu2026dcut}. For MoE targets, Cascade adapts verification budgets~\citep{saxena2025cascade}, EVICT truncates draft trees using drafter signals and profiled verification costs~\citep{pan2026evict}, and MoESD studies speculative decoding for sparse models~\citep{huang2025moesd}. EcoSpec prioritizes draft-tree nodes likely to reuse experts using a small model that predicts the target's expert assignments~\citep{xie2026ecospec}. FASER, designed for continuous-batching serving, prunes tokens predicted to be rejected at intermediate target layers~\citep{chen2026faser}.

\subsection{Gated DeltaNet and chunkwise verification}
\label{sec:gdnbackground}
\label{sec:recurrentrelated}

GDN is a linear-attention layer that keeps a fixed-size state matrix as a memory from keys to values~\citep{katharopoulos2020linear,yang2025gated}. In the evaluated models, GDN layers are interleaved with softmax-attention layers, and every layer ends with an MoE block. For one head at token position $i$, let $q_i,k_i\in\R^{d_k}$ be the query and key, $v_i\in\R^{d_v}$ the value, and $S_i\in\R^{d_v\times d_k}$ the state after token $i$. Linear projections followed by a short causal convolution produce $q_i$, $k_i$, and $v_i$, and $q_i$ and $k_i$ are normalized to unit length. Each token decays the state by $\alpha_i\in(0,1)$ and then corrects the value stored under its own key with a writing strength $\beta_i\in(0,1)$. Both come from linear projections, and GDN applies
\begin{equation}
 z_i=\beta_i\big(v_i-\alpha_i S_{i-1}k_i\big),
 \qquad S_i=\alpha_i S_{i-1}+z_i k_i^\top.
 \label{eq:gdn}
\end{equation}
The update $z_i\in\R^{d_v}$ moves the value stored under $k_i$ toward $v_i$ by a fraction $\beta_i$, which is the delta rule~\citep{schlag2021linear}. The output $o_i=S_iq_i$ reads the updated state with the query, and we omit later output transformations such as normalization and the output gate.

Autoregressive decoding applies Equation~\eqref{eq:gdn} one token at a time. A verifier that does the same must store $S_i$, a $d_v\times d_k$ matrix per head, after every token, so that it can later restore the state at the end of whichever prefix is accepted. Verification can avoid this cost because all tokens of the target input, the tokens that the target processes in one round, are known in advance. Chunkwise formulations, developed for efficient training, process the tokens of a chunk in parallel~\citep{dao2024ssm,yang2024parallel}. Number the tokens of the target input $1,\ldots,w$, and let $S_0$ be the state before them. With the cumulative log-decay $g_i=\sum_{j=1}^{i}\log\alpha_j$, unrolling Equation~\eqref{eq:gdn} from $S_0$ gives:
\begin{equation}
 S_i=e^{g_i}S_0+\sum_{j=1}^{i}e^{g_i-g_j}z_jk_j^\top.
 \label{eq:unrolled}
\end{equation}
Substituting this expression for $S_{i-1}$ into Equation~\eqref{eq:gdn} and stacking $k_i^\top$, $v_i^\top$, and $z_i^\top$ as the rows of $K\in\R^{w\times d_k}$ and $V,Z\in\R^{w\times d_v}$ gives one lower-triangular linear system for all $w$ updates,
\begin{equation}
 HZ=\operatorname{diag}(\beta)\big(V-\operatorname{diag}(e^{g})KS_0^\top\big),
 \label{eq:blockoutputs}
\end{equation}
where $\beta=(\beta_1,\ldots,\beta_w)$, $e^{g}=(e^{g_1},\ldots,e^{g_w})$, and $H$ has ones on its diagonal and entries $H_{ij}=\beta_ie^{g_i-g_j}k_i^\top k_j$ for $j<i$. Solving this small $w\times w$ triangular system once therefore replaces the $w$ sequential state updates. The solve and the dense matrix products that then give all $w$ outputs use only $S_0$ and matrices with $w$ rows, so no intermediate $d_v\times d_k$ state is formed. We call $(z_i,k_i,g_i)$ the compact factors of token $i$, because Equation~\eqref{eq:unrolled} reconstructs any state $S_i$ from $S_0$ and the factors of tokens $1,\ldots,i$. \method{} uses them to reconstruct the state after the accepted prefix.

Speculative decoders for recurrent targets differ in how they recover the state after the accepted prefix. STree accumulates state transitions to verify token trees~\citep{wu2025stree}. SpecLA, Bole, and TreeWY retain compact factors from which the accepted state is reconstructed~\citep{wang2026specla,wang2026bole,ghantasala2026treewy}. Other hybrid and recurrent systems likewise avoid storing a state for every verified token~\citep{wang2024mamballama,wu2024snakes}. Concurrent work, ReplaySSM, verifies drafts with a chunkwise triangular solve and writes the state back only when its ring buffer of recent GDN updates is flushed~\citep{liou2026replayssm}. Its reference implementation in vLLM~\citep{kwon2023vllm} targets large serving batch sizes, whereas \method{} targets single-request decoding. Equation~\eqref{eq:blockoutputs} is the standard chunkwise computation. \method{} contributes fused, block-parallel verification kernels and a commit kernel that reconstructs the accepted state of every GDN layer in one launch.

\section{HybridSpec}
\label{sec:setting}
\label{sec:method}
\method{} selects the verification length entirely on the GPU and splits target verification at the router boundary: immediately after the MoE router in the first layer (Layer~0) assigns experts, but before any routed expert is executed. Figure~\ref{fig:hybridspec-overview} illustrates one decoding round, and Algorithm~\ref{alg:hybridspec} in Appendix~\ref{app:roundprocedure} details the complete procedure.

\begin{figure}[t]
\centering
\includegraphics[width=\linewidth]{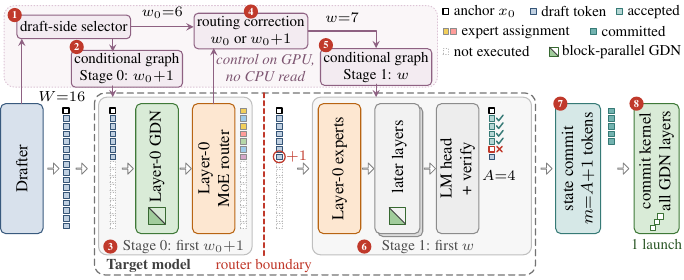}
\caption{One round of \method{} in single-request greedy decoding, with illustrative values: nominal verification length $w_0=6$, executed verification length $w=7$, and $A=4$ accepted draft tokens. Circled numbers mark the steps described in Section~\ref{sec:round}.}
\label{fig:hybridspec-overview}
\vspace{-15pt}
\end{figure}

\subsection{Setting and throughput objective}
\label{sec:semantics}
\label{sec:model}

We first formulate \method{} for greedy decoding with sequence batch size $B=1$. In each round, a fixed drafter produces a proposal block of $D$ draft tokens. Together with the anchor, the newest emitted token that the target has not yet processed, this block forms a full target input of length $W=D+1$. \method{} retains the first $M\leq D$ draft tokens, which we call the~\textit{proposal prefix}, and verifies the corresponding target input of length $w=M+1$. Thus $D$ and $M$ count draft tokens, whereas $W$ and $w$ count target input tokens, including the anchor. In all evaluated models, Layer~0 applies a GDN layer followed by an MoE block.

At the start of a round, $x_0$ denotes the anchor and $d_i$ the $i$-th draft token. To make the first $M$ draft tokens eligible for acceptance, the target processes $(x_0,d_1,\ldots,d_M)$. If a nonterminal round accepts $A\leq M$ draft tokens, it emits $R=A+1$ tokens and commits $m=A+1$ input tokens. These counts are equal but refer to different sequences: the emitted tokens are $(d_1,\ldots,d_A,x_{\mathrm{new}})$, whereas the committed input tokens are $(x_0,d_1,\ldots,d_A)$. Here $x_{\mathrm{new}}$ is the token that the target emits after the accepted prefix. The old anchor is therefore consumed even when $A=0$, and $x_{\mathrm{new}}$ becomes the next anchor. Both sequences are truncated at an end-of-sequence (EOS) token or the output limit.

Verifying only a prefix must not change the output. Under greedy decoding, \method{} emits the same tokens as target-only decoding in exact arithmetic, as Proposition~\ref{prop:greedy} in Appendix~\ref{app:greedy} proves under the assumptions stated there. Under temperature sampling, the verifier accepts each draft token with the probability that the target assigns to it and, after a rejection, samples a replacement from the target's distribution with that token excluded. Any verification length fixed before these draws, including the length after the routing correction, therefore preserves the output distribution of target-only sampling in exact arithmetic, as Proposition~\ref{prop:sampling} proves.

Let $\mathcal I$ denote the information available when selecting a verification length and $A_D$ the number of draft tokens accepted from the full proposal block. A proposal prefix of length $M$ has $\min(A_D,M)$ accepted draft tokens, and the round emits one more token. The prefix-acceptance probability $p_i(\mathcal I)=\Prb(A_D\geq i\mid\mathcal I)$ is the probability that the first $i$ draft tokens are accepted. For a nonterminal round, the expected number of emitted tokens is ${G(M\mid\mathcal I)}=1+\sum_{i=1}^{M}p_i(\mathcal I)$. We compare all values of $M$ at the same initial state and proposal block. Let $L(M\mid\mathcal I)$ denote the expected latency of a complete round, including drafting, all target computation, length-selection and graph-launch overhead, acceptance, and state commit. The objective is pooled decode throughput, the total number of emitted tokens divided by the total decode time. For a fixed set of rounds with information $\mathcal I_1,\ldots,\mathcal I_N$, choosing $M_r$ in round $r$ gives the pooled throughput
\begin{equation}
 \frac{\sum_{r=1}^{N}G(M_r\mid\mathcal I_r)}{\sum_{r=1}^{N}L(M_r\mid\mathcal I_r)}.
 \label{eq:pooled}
\end{equation}
This ratio of sums is not the average of the per-round ratios $G/L$, so the best choice in one round depends on the throughput of all rounds. Let $\rho^\star$ be the maximum of Equation~\eqref{eq:pooled}. By Dinkelbach's theorem~\citep{dinkelbach1967fractional}, a set of choices attains $\rho^\star$ exactly when every round maximizes ${G(M_r\mid\mathcal I_r)}-\rho^\star {L(M_r\mid\mathcal I_r)}$. Appendix~\ref{app:width} analyzes this per-round score.

\subsection{Two-stage verification at the router boundary}
\label{sec:overview}
\label{sec:graphruntime}
\label{sec:round}

After drafting, the draft-side selector chooses the nominal verification length $w_0$ from the drafter's signals and writes it to GPU memory (step~1 in Figure~\ref{fig:hybridspec-overview}). Let $w_{\mathrm{rem}}\in\{2,\ldots,W\}$ be the largest verification length that the remaining output budget allows, but at least 2. Stage~0 covers the first $\min(w_0+1,w_{\mathrm{rem}})$ tokens of the target input and stops at the router boundary, before any routed expert runs. \method{} captures in advance one CUDA Graph, a recorded sequence of GPU kernels, for each possible Stage-0 length. A conditional graph node~\citep{nvidia2026cudagraphs} reads the Stage-0 length and launches the matching graph without a CPU read of the length (step~2). Stage~0 then runs these tokens through the Layer-0 GDN layer and MoE router (step~3). It retains the hidden states and residuals needed to resume the layer, plus each token's expert assignments and routing weights.

At the router boundary, the routing correction decides whether to keep $d_{w_0}$, the extra draft token that Stage~0 processed beyond the first $w_0$ tokens (step~4). The draft-side selector chose $w_0$ before the target ran, so it could charge $d_{w_0}$ only the average expert cost. The actual cost is lower if $d_{w_0}$ reuses experts already selected by the first $w_0$ tokens and higher if it activates new ones. Because Stage~0 has already routed $d_{w_0}$, the correction counts the new Layer-0 experts that it adds and uses this count to predict its expert cost in all MoE layers. It keeps $d_{w_0}$, setting the corrected verification length to $w_{\mathrm c}=w_0+1$, only if its acceptance probability exceeds the predicted latency that it adds, converted into tokens at the pooled throughput rate, and otherwise sets $w_{\mathrm c}=w_0$, as Equation~\eqref{eq:routing-correction} formalizes. The executed verification length is $w=\min(w_{\mathrm c},w_{\mathrm{rem}})$. Stage~1 then processes the first $w$ tokens from the Layer-0 experts through the final layer without repeating Stage~0. \method{} likewise captures one Stage-1 graph for each possible value of $w$, and a second conditional graph launches the graph for $w$ (step~5). This graph reads the retained values of the first $w$ tokens, builds the attention and GDN metadata for length $w$, and executes the final normalization and projection and the greedy verification of the $w-1$ draft tokens (step~6).

Once verification accepts $A$ draft tokens, the Stage-1 graph ends with the commit, which advances the attention and convolution caches, as well as the target hidden states that the drafter reads, through the $m=A+1$ committed tokens (step~7). Layer-0 GDN retains compact factors for the Stage-0 tokens, and later GDN layers retain factors for the $w$ selected tokens. A commit kernel then rebuilds the accepted state of every GDN layer from these factors in one launch (step~8).

\subsection{Draft-side selection and routing correction}
\label{sec:benefit}
\label{sec:routingselection}

The length policy has two parts: the draft-side selector, which makes the main decision before target execution, and a lightweight routing correction, which runs after Stage~0 and can add one token. The fixed drafter supplies a top-1 confidence and an entropy for each of its ${D=15}$ draft tokens. Before Layer~0, the selector calibrates the confidence and computes an entropy-change feature within the proposal block. It estimates conditional acceptance probabilities $\hat c_i\approx\Prb(A_D\geq i\mid A_D\geq i-1,\mathcal I)$ and prefix probabilities $\hat p_i=\prod_{j=1}^{i}\hat c_j$, so ${1\geq\hat p_1\geq\cdots\geq\hat p_D\geq0}$. In terms of verification length $w=M+1$, the expected number of emitted tokens $G$ is $1+\sum_{i=1}^{w-1}p_i$. Over the nominal verification lengths $\mathcal W_0=\{2,\ldots,W\}$, the selector chooses
\begin{equation}
 w_0=\argmax_{w\in\mathcal W_0}
 \left[1+\sum_{i=1}^{w-1}\hat p_i-\rho C_{\mathrm{tgt}}(w)\right],
 \label{eq:nominal-selector}
\end{equation}
where the target-cost curve $C_{\mathrm{tgt}}(w)$ is the measured target latency at length $w$ and $\rho>0$ is a calibrated estimate of $\rho^\star$ that converts latency into tokens. The score is the per-round Dinkelbach objective with estimated acceptance probabilities. It uses $C_{\mathrm{tgt}}$ in place of $L$, because length-independent costs do not change its maximizer. Ties favor the shorter target input.

\begin{figure}[!t]
\centering
\includegraphics[width=\linewidth]{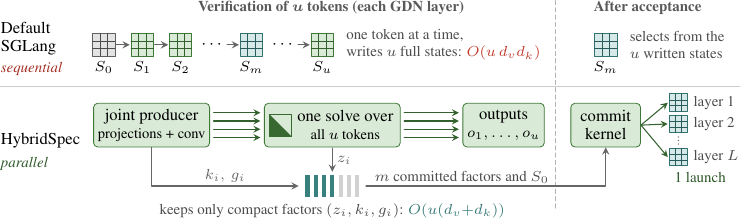}
\caption{GDN verification and accepted-state reconstruction in the default SGLang verifier and in \method{}. The default verifier steps through the $u$ tokens one at a time, whereas \method{} processes all $u$ tokens in parallel and rebuilds the accepted states of all GDN layers in parallel, in one launch. Teal marks the accepted state and the factors of the $m$ committed tokens.}
\label{fig:gdn-kernels}
\vspace{-15pt}
\end{figure}

The routing correction uses the expert assignments observed in Stage~0 only to revise the latency estimate, and it leaves the acceptance probabilities $\hat p_i$ unchanged. Let $U_0(w)$ be the number of distinct Layer-0 experts selected by the first $w$ tokens of the target input. Extending the target input from $w_0$ to $w_0+1$ adds draft token $d_{w_0}$ and raises the expected number of emitted tokens by $\hat p_{w_0}$. Because drafting and Stage~0 are complete when the correction runs, the extension changes only the latency of Stage~1. The curve increment $C_{\mathrm{tgt}}(w_0+1)-C_{\mathrm{tgt}}(w_0)$ in Equation~\eqref{eq:nominal-selector} differs from this change in two ways that the correction can account for. First, it includes the cost $s$ of running one more token through the Layer-0 GDN layer and router, which Stage~0 has already paid for $d_{w_0}$. Second, it charges expert time at its calibration average. At batch size one, each expert processes only a few tokens, so MoE latency depends mainly on the number of distinct experts. Each distinct (layer, expert) pair therefore adds about $\eta$ to the latency. Let $n_{\mathrm{all}}$ be the number of new (layer, expert) pairs that $d_{w_0}$ adds across all MoE layers. The curve thus charges about $\eta\,\bar n_{\mathrm{all}}(w_0)$ for them, where $\bar n_{\mathrm{all}}(w_0)$ is the average of $n_{\mathrm{all}}$ at this position. Stage~0 observes only the Layer-0 part of $n_{\mathrm{all}}$, $n_{\mathrm{new}}=U_0(w_0+1)-U_0(w_0)$. Because later-layer routing correlates with Layer-0 routing, as Appendix~\ref{app:widthprofile} shows, the correction replaces $\bar n_{\mathrm{all}}(w_0)$ with $\bar n_{\mathrm{all}}(w_0\mid n_{\mathrm{new}})$, the average of $n_{\mathrm{all}}$ given $n_{\mathrm{new}}$. Subtracting $s$ from the curve increment and making this replacement gives $\Delta L_{\mathrm c}$, the estimated Stage-1 latency increase. The correction applies the additive score of Equation~\eqref{eq:nominal-selector}, with the same rate $\rho$, to this one-step choice:
\begin{align}
 \Delta L_{\mathrm c}&=C_{\mathrm{tgt}}(w_0+1)-C_{\mathrm{tgt}}(w_0)-s+\eta\big(\bar n_{\mathrm{all}}(w_0\mid n_{\mathrm{new}})-\bar n_{\mathrm{all}}(w_0)\big),\nonumber\\
 w_{\mathrm c}&=
 \begin{cases}
 w_0+1, & \text{if } \hat p_{w_0}>\rho\,\Delta L_{\mathrm c},\\
 w_0, & \text{otherwise}.
 \end{cases}
 \label{eq:routing-correction}
\end{align}
Because $w_0$ maximizes the score in Equation~\eqref{eq:nominal-selector} under the same curve, $\hat p_{w_0}\leq\rho\,[C_{\mathrm{tgt}}(w_0+1)-C_{\mathrm{tgt}}(w_0)]$ whenever $w_0<W$. The correction can therefore expand only if $d_{w_0}$ costs less than the curve predicts, that is, only if $s>\eta\,\big(\bar n_{\mathrm{all}}(w_0\mid n_{\mathrm{new}})-\bar n_{\mathrm{all}}(w_0)\big)$.

The correction adds at most one token because Stage~0 must process every token that the correction might add, at cost $s$ per token in every round. If $w_0=W$ or $w_{\mathrm{rem}}\leq w_0$, Stage~0 does not cover position $w_0+1$, and the correction does not expand. Acceptance parameters are fitted on the calibration prompts, latency is measured on the evaluated runtime, and all parameters are frozen before test evaluation. Appendix~\ref{app:costfeatures} describes the calibration.

\subsection{Parallel GDN verification and accepted-state reconstruction}
\label{sec:gdnmethod}

When a GDN layer processes $u$ tokens, the default SGLang verifier updates its state one token at a time and writes a full state after every token, requiring $O(u\,d_vd_k)$ elements per head. \method{} instead treats the target input as a single chunk and processes all $u$ tokens in parallel, as Figure~\ref{fig:gdn-kernels} shows. Its block-parallel triangular-solve kernel solves Equation~\eqref{eq:blockoutputs}, computes the outputs from the result, and retains only the compact factors $(z_i,k_i,g_i)$, which require $O(u(d_v+d_k))$ elements. Both counts exclude $S_0$ and workspace. The Layer-0 GDN layer precedes the router, so it runs on the \mbox{Stage-0} tokens with $u=\min(w_0+1,w_{\mathrm{rem}})$, and later GDN layers run with $u=w$ in Stage~1. For $2\leq w\leq16$, the kernels are compiled for each length, so they evaluate only the selected tokens, and exploit the small size of $H$ with closed-form $4\times4$ block inverses and an exact truncated series. A joint producer computes all input projections and the convolution in one launch.

After verification, each GDN layer must advance its state through the $m=A+1$ committed tokens, namely the anchor and the accepted draft tokens, as the right side of Figure~\ref{fig:gdn-kernels} shows. Let $Z_{1:m}$ and $K_{1:m}$ denote the first $m$ rows of $Z$ and $K$ in Equation~\eqref{eq:blockoutputs}. With $S_0$ the layer's state at the start of the round, the accepted state is Equation~\eqref{eq:unrolled} at $i=m$. In matrix form, it reads
\begin{equation}
 S_m=e^{g_m}S_0+Z_{1:m}^\top\operatorname{diag}\big(e^{g_m-g_1},\ldots,e^{g_m-g_m}\big)K_{1:m}.
 \label{eq:commit}
\end{equation}
Because $H$ is lower triangular, the first $m$ rows of $Z$ depend only on the first $m$ tokens. Discarding the rejected suffix therefore leaves $Z_{1:m}$ unchanged in exact arithmetic, so the commit kernel reuses the factors computed during verification. The \method{} kernels store $k_i$ and $g_i$ rather than decay-scaled keys, so the commit kernel can form the decay weights $e^{g_m-g_i}$ for any $m$. Accepted-state reconstruction is independent across GDN layers, so one commit kernel launch reconstructs the state of every GDN layer. The kernel accumulates in 32-bit floating point (FP32). Only this reconstruction is parallel across layers, and the target forward pass remains sequential.

\section{Evaluation}
\label{sec:evaluation}

We first compare \method{} end to end against six baselines across three target--drafter pairs. We then use matched ablations and component-level measurements to isolate the contributions of adaptive length selection, routing correction, and the GDN kernels.

\subsection{Experimental setup}
\label{sec:setup}

\paragraph{Models and workloads.}
We evaluate Qwen3.5-35B-A3B, Qwen3.6-35B-A3B, and Qwen3.5-122B-A10B-GPTQ-INT4~\citep{qwen2026qwen35a3b,qwen2026qwen36a3b,qwen2026qwen35a10b}, with their official DFlash drafters, in SGLang 0.5.16 with one request at a time on NVIDIA A100 80~GB GPUs. Prompts come from seven categories of the SPEED-Bench Qualitative split~\citep{abramovich2026speedbench}: coding, math, reasoning, retrieval-augmented generation (RAG), summarization, writing, and roleplay. A seeded split gives 280 calibration and 280 test prompts, 40 per category in each. Table~\ref{tab:main-greedy} reports greedy decoding, the primary setting, and Appendices~\ref{app:sampling-results} and~\ref{app:batching} report temperature-one sampling and batched decoding.

\paragraph{Baselines.}
We compare \method{} with autoregressive (AR) decoding, vanilla DFlash, \mbox{D-Cut}~\citep{liu2026dcut}, EVICT~\citep{pan2026evict}, ReplaySSM~\citep{liou2026replayssm}, and EVICT + ReplaySSM. All speculative baselines use the same DFlash drafter as \method{}. We adapt \mbox{D-Cut} and EVICT to the DFlash proposal block and use the SGLang implementation of ReplaySSM. Vanilla DFlash verifies the full target input of length $W=D+1$ in every round. \mbox{D-Cut} and EVICT shorten the proposal block before verification. ReplaySSM replaces the default GDN verification procedure, and EVICT + ReplaySSM combines both optimizations.

\paragraph{Metrics.}
The primary metric is the speedup ratio (SR), defined as the pooled post-prefill decode throughput normalized by that of the corresponding AR baseline. Acceptance length (AL) is the number of tokens emitted per verification round, including the accepted draft tokens and the additional token generated by the target. Appendix~\ref{app:e2eprotocol} details the full protocol.


\subsection{End-to-end performance}

\begin{table}[h]
\centering\scriptsize
\setlength{\tabcolsep}{1.9pt}
\caption{End-to-end comparison under greedy decoding. AL is the number of tokens emitted per verification round, and SR is the decode-throughput speedup over paired autoregressive (AR) decoding. Summ.\ denotes summarization, and Avg.\ is the mean over the seven categories.}
\label{tab:main-greedy}
\begin{tabular}{@{}lcccccccccccccccc@{}}
\toprule
 & \multicolumn{2}{c}{Coding} & \multicolumn{2}{c}{Math} & \multicolumn{2}{c}{Reasoning} & \multicolumn{2}{c}{RAG} & \multicolumn{2}{c}{Summ.} & \multicolumn{2}{c}{Writing} & \multicolumn{2}{c}{Roleplay} & \multicolumn{2}{c}{Avg.} \\
\cmidrule(lr){2-3}\cmidrule(lr){4-5}\cmidrule(lr){6-7}\cmidrule(lr){8-9}\cmidrule(lr){10-11}\cmidrule(lr){12-13}\cmidrule(lr){14-15}\cmidrule(lr){16-17}
Method & AL & SR & AL & SR & AL & SR & AL & SR & AL & SR & AL & SR & AL & SR & AL & SR \\
\midrule
\multicolumn{17}{@{}l}{\textit{Qwen3.5-35B-A3B}} \\
AR target            & \notapplicable & $1.00\times$ & \notapplicable & $1.00\times$ & \notapplicable & $1.00\times$ & \notapplicable & $1.00\times$ & \notapplicable & $1.00\times$ & \notapplicable & $1.00\times$ & \notapplicable & $1.00\times$ & \notapplicable & $1.00\times$ \\
Vanilla DFlash       & 6.43 & $1.89\times$ & 5.40 & $1.54\times$ & 4.91 & $1.43\times$ & 6.07 & $1.72\times$ & 4.96 & $1.38\times$ & 3.92 & $1.13\times$ & 3.53 & $1.05\times$ & 5.03 & $1.45\times$ \\
\mbox{D-Cut}         & 6.12 & $1.98\times$ & 4.97 & $1.64\times$ & 4.61 & $1.54\times$ & 5.76 & $1.80\times$ & 4.65 & $1.48\times$ & 3.70 & $1.26\times$ & 3.37 & $1.19\times$ & 4.74 & $1.55\times$ \\
EVICT                & 6.18 & $1.99\times$ & 5.15 & $1.69\times$ & 4.68 & $1.58\times$ & 5.88 & $1.85\times$ & 4.78 & $1.54\times$ & 3.77 & $1.29\times$ & 3.36 & $1.20\times$ & 4.83 & $1.59\times$ \\
ReplaySSM            & 6.41 & $1.82\times$ & 5.53 & $1.54\times$ & 4.82 & $1.36\times$ & 6.04 & $1.66\times$ & 4.93 & $1.34\times$ & 3.92 & $1.10\times$ & 3.51 & $1.03\times$ & 5.02 & $1.41\times$ \\
EVICT + ReplaySSM    & 5.94 & $1.91\times$ & 5.11 & $1.65\times$ & 4.60 & $1.52\times$ & 5.69 & $1.76\times$ & 4.72 & $1.50\times$ & 3.73 & $1.26\times$ & 3.33 & $1.17\times$ & 4.73 & $1.54\times$ \\
\oursrow \method{}            & 6.20 & $2.39\times$ & 5.10 & $1.97\times$ & 4.62 & $1.87\times$ & 5.87 & $2.18\times$ & 4.80 & $1.84\times$ & 3.69 & $1.51\times$ & 3.27 & $1.42\times$ & 4.79 & $1.88\times$ \\
\midrule
\multicolumn{17}{@{}l}{\textit{Qwen3.6-35B-A3B}} \\
AR target            & \notapplicable & $1.00\times$ & \notapplicable & $1.00\times$ & \notapplicable & $1.00\times$ & \notapplicable & $1.00\times$ & \notapplicable & $1.00\times$ & \notapplicable & $1.00\times$ & \notapplicable & $1.00\times$ & \notapplicable & $1.00\times$ \\
Vanilla DFlash       & 5.58 & $1.56\times$ & 5.41 & $1.58\times$ & 4.79 & $1.42\times$ & 5.87 & $1.63\times$ & 4.85 & $1.37\times$ & 3.66 & $1.06\times$ & 3.33 & $0.99\times$ & 4.79 & $1.37\times$ \\
\mbox{D-Cut}         & 5.36 & $1.66\times$ & 5.11 & $1.68\times$ & 4.58 & $1.54\times$ & 5.62 & $1.71\times$ & 4.67 & $1.50\times$ & 3.53 & $1.19\times$ & 3.21 & $1.12\times$ & 4.58 & $1.48\times$ \\
EVICT                & 5.37 & $1.68\times$ & 5.17 & $1.70\times$ & 4.63 & $1.58\times$ & 5.63 & $1.73\times$ & 4.72 & $1.54\times$ & 3.52 & $1.21\times$ & 3.17 & $1.13\times$ & 4.60 & $1.51\times$ \\
ReplaySSM            & 5.65 & $1.54\times$ & 5.49 & $1.55\times$ & 4.83 & $1.39\times$ & 5.69 & $1.53\times$ & 4.87 & $1.34\times$ & 3.66 & $1.04\times$ & 3.36 & $0.98\times$ & 4.79 & $1.34\times$ \\
EVICT + ReplaySSM    & 5.37 & $1.63\times$ & 5.18 & $1.65\times$ & 4.60 & $1.53\times$ & 5.59 & $1.66\times$ & 4.69 & $1.48\times$ & 3.54 & $1.18\times$ & 3.19 & $1.10\times$ & 4.59 & $1.46\times$ \\
\oursrow \method{}            & 5.28 & $2.05\times$ & 5.14 & $2.09\times$ & 4.62 & $1.88\times$ & 5.55 & $2.07\times$ & 4.74 & $1.85\times$ & 3.50 & $1.46\times$ & 3.16 & $1.39\times$ & 4.57 & $1.83\times$ \\
\midrule
\multicolumn{17}{@{}l}{\textit{Qwen3.5-122B-A10B}} \\
AR target            & \notapplicable & $1.00\times$ & \notapplicable & $1.00\times$ & \notapplicable & $1.00\times$ & \notapplicable & $1.00\times$ & \notapplicable & $1.00\times$ & \notapplicable & $1.00\times$ & \notapplicable & $1.00\times$ & \notapplicable & $1.00\times$ \\
Vanilla DFlash       & 6.83 & $1.67\times$ & 5.60 & $1.41\times$ & 5.13 & $1.29\times$ & 6.31 & $1.56\times$ & 5.33 & $1.31\times$ & 4.13 & $1.05\times$ & 3.54 & $0.95\times$ & 5.27 & $1.32\times$ \\
\mbox{D-Cut}         & 6.29 & $1.89\times$ & 5.23 & $1.70\times$ & 4.85 & $1.59\times$ & 5.95 & $1.81\times$ & 5.01 & $1.60\times$ & 3.85 & $1.32\times$ & 3.44 & $1.24\times$ & 4.95 & $1.59\times$ \\
EVICT                & 6.05 & $1.93\times$ & 5.09 & $1.76\times$ & 4.79 & $1.66\times$ & 5.94 & $1.90\times$ & 4.82 & $1.63\times$ & 3.72 & $1.37\times$ & 3.33 & $1.29\times$ & 4.82 & $1.65\times$ \\
ReplaySSM            & 6.59 & $1.56\times$ & 5.78 & $1.42\times$ & 5.08 & $1.25\times$ & 6.37 & $1.53\times$ & 5.26 & $1.28\times$ & 4.06 & $1.02\times$ & 3.60 & $0.95\times$ & 5.25 & $1.29\times$ \\
EVICT + ReplaySSM    & 6.31 & $1.89\times$ & 5.10 & $1.68\times$ & 4.70 & $1.56\times$ & 5.99 & $1.82\times$ & 4.97 & $1.60\times$ & 3.78 & $1.32\times$ & 3.38 & $1.25\times$ & 4.89 & $1.59\times$ \\
\oursrow \method{}            & 6.02 & $2.06\times$ & 5.10 & $1.84\times$ & 4.71 & $1.77\times$ & 5.85 & $2.01\times$ & 4.91 & $1.72\times$ & 3.75 & $1.46\times$ & 3.34 & $1.38\times$ & 4.81 & $1.75\times$ \\
\bottomrule
\end{tabular}
\vspace{-15pt}
\end{table}

In Table~\ref{tab:main-greedy}, the mean SR of \method{} is $1.88\times$ on Qwen3.5-35B-A3B, $1.83\times$ on Qwen3.6-35B-A3B, and $1.75\times$ on Qwen3.5-122B-A10B. Vanilla DFlash reaches $1.45\times$, $1.37\times$, and $1.32\times$, and EVICT, the strongest baseline on all three pairs, reaches $1.59\times$, $1.51\times$, and $1.65\times$. \method{} has the highest SR in every category, and its per-category SR ranges from $1.38\times$ for roleplay on Qwen3.5-122B-A10B to $2.39\times$ for coding on Qwen3.5-35B-A3B. And we further evaluate correctness and task quality in Appendix~\ref{sec:correctnessresults}.

\begin{wrapfigure}{r}{0.46\linewidth}
\vspace{-10pt}
\centering
\includegraphics[width=\linewidth]{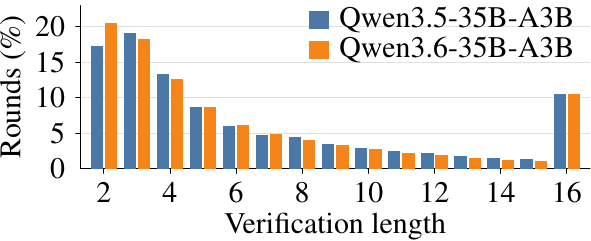}
\vspace{-2.0em}
\caption{\raggedright Executed verification lengths of \method{} on the 280 test prompts. Vanilla DFlash verifies 16 tokens per round.}
\label{fig:length-histogram}
\vspace{-0.7em}
\end{wrapfigure}

\method{} has a lower AL than vanilla DFlash but a higher SR mainly because it verifies fewer tokens in each round. On the two 35B pairs, the mean verification length of \method{} is 6.4 and 6.2, compared with 16 for vanilla DFlash, and Figure~\ref{fig:length-histogram} shows that about half of its rounds verify at most 4 tokens. Relative to vanilla DFlash, \method{} lowers mean AL by 0.24 on Qwen3.5-35B-A3B and 0.22 on Qwen3.6-35B-A3B, or 5\% on both. It leaves 9.6 and 9.8 draft tokens per round unverified, and the AL difference corresponds to fewer than 3\% of them. Most skipped tokens would therefore have been rejected. Meanwhile, the decode time per round, the pooled decode time divided by the number of rounds, falls from 22.9 to 16.5~ms on Qwen3.5-35B-A3B and from 23.0 to 16.3~ms on Qwen3.6-35B-A3B. In every category on both 35B pairs, AL falls by at most 7\%, and time per round falls by 24\%--32\%. Time per round falls most for writing and roleplay, the two categories with the lowest AL. In the isolated benchmarks of Table~\ref{tab:gdnprelim}, with every draft token accepted, the GDN verification and reconstruction kernels save 0.12~ms per round at length 2 and 1.44~ms at length 16.

\mbox{D-Cut} and EVICT reach an AL close to that of \method{} (4.58--4.95 versus 4.57--4.81) but a lower SR (1.48$\times$--1.65$\times$ versus 1.75$\times$--1.88$\times$), so \method{} spends less time per round. Four design differences can contribute. First, both baselines choose the length that maximizes a per-round ratio of expected tokens to profiled step latency~\citep{liu2026dcut,pan2026evict}, whereas \method{} maximizes pooled throughput through the per-round score of Equation~\eqref{eq:nominal-selector}. Second, both form acceptance estimates directly from drafter confidences, whereas \method{} calibrates its estimates on the calibration prompts. Third, both read the selected length on the CPU to launch the matching verification graph, whereas \method{} launches it with a conditional graph node and never copies the length to the CPU. Fourth, \method{} uses its own GDN kernels. EVICT + ReplaySSM, which adds prior GDN verification to EVICT, still trails \method{} by 0.16--0.37 in mean SR.
Without recalibration, \method{} also speeds up the four other SPEED-Bench Qualitative categories. In Table~\ref{tab:ood}, where DFlash denotes vanilla DFlash, \method{} reaches a mean SR of $1.69\times$ on their 320 prompts with Qwen3.6-35B-A3B, versus $1.31\times$ for vanilla DFlash, and leads in every category.

\begin{table}[H]
\centering\footnotesize
\setlength{\tabcolsep}{2.8pt}
\setlength{\aboverulesep}{0pt}
\setlength{\belowrulesep}{0pt}
\renewcommand{\arraystretch}{1.25}
\begin{minipage}[t]{0.37\linewidth}
\centering
\caption{\raggedright Unseen SPEED-Bench categories on \mbox{Qwen3.6-35B-A3B}, evaluated without recalibration.\strut}
\label{tab:ood}
\begin{tabular}{@{}lcc>{\columncolor{oursrowbg}}c>{\columncolor{oursrowbg}[\tabcolsep][0pt]}c@{}}
\toprule
 & \multicolumn{2}{c}{DFlash} & \multicolumn{2}{>{\columncolor{oursrowbg}[\tabcolsep][0pt]}c}{\method{}} \\
\cmidrule(lr){2-3}\cmidrule(l){4-5}
Category & AL & SR & AL & SR \\
\midrule
Humanities   & 3.89 & $1.18\times$ & 3.65 & $1.55\times$ \\
QA           & 3.95 & $1.19\times$ & 3.64 & $1.56\times$ \\
STEM         & 4.27 & $1.28\times$ & 3.95 & $1.66\times$ \\
Multilingual & 5.27 & $1.57\times$ & 4.98 & $2.01\times$ \\
\midrule
Avg.         & 4.35 & $1.31\times$ & 4.06 & $1.69\times$ \\
\bottomrule
\end{tabular}
\end{minipage}\hfill
\begin{minipage}[t]{0.59\linewidth}
\centering
\caption{\raggedright Component gains on \mbox{Qwen3.6-35B-A3B}. Gain is the SR of \method{} divided by that of the variant, and $w$ is the mean verification length.\strut}
\label{tab:ablations}
\begin{tabular}{@{}llcccc@{}}
\toprule
Component & Replaced by & $w$ & AL & SR & Gain \\
\midrule
\oursrow \method{} & -- & 6.2 & 4.57 & $1.83\times$ & -- \\
Length selection & Static length 7 & 7 & 3.99 & $1.54\times$ & $1.19\times$ \\
Length selection & Length 16 & 16 & 4.77 & $1.46\times$ & $1.25\times$ \\
Routing correction & Nominal-only & 6.1 & 4.48 & $1.73\times$ & $1.06\times$ \\
GDN kernels & SGLang kernels & 6.2 & 4.51 & $1.74\times$ & $1.05\times$ \\
\midrule
All components & Vanilla DFlash & 16 & 4.79 & $1.37\times$ & $1.33\times$ \\
\bottomrule
\end{tabular}
\end{minipage}\
\vspace{-10pt}
\end{table}

\subsection{Ablations}
\label{sec:ablations}
Table~\ref{tab:ablations} removes one component of \method{} at a time on Qwen3.6-35B-A3B while keeping the others unchanged. We report the gain as the SR of \method{} divided by that of each variant, representing the speedup contributed by the removed component relative to its stated replacement. Vanilla DFlash removes all components. Removing length selection also eliminates two-stage verification and the routing correction, since both depend on the selected length. We therefore evaluate two fixed-length variants that use the \method{} GDN kernels within a single CUDA Graph. The calibrated static length is 7, the fixed length with the highest pooled throughput on the calibration prompts among screened candidates, while length 16 matches vanilla DFlash. The nominal-only control removes only the routing correction: it keeps $w=w_0$ under the same two-stage verification, with Stage~0 processing only the first $w_0$ tokens. Thus, its measured gain captures the routing correction net of the cost of processing the extra Stage-0 token. Finally, the GDN variant retains the full length-selection policy but replaces the \method{} GDN kernels with the default SGLang kernels.

Length selection provides the largest gain: $1.19\times$ over the calibrated static length and $1.25\times$ over length~16. Although \method{} verifies fewer tokens per round than the static-length variant, 6.2 versus 7, it achieves a higher AL, 4.57 versus 3.99, by adapting the verification length to each round. The routing correction provides a further $1.06\times$ gain over the nominal-only control. The \method{} GDN kernels contribute $1.05\times$ at the lengths selected by \method{}. At length~16, the SR ratio between the fixed-length variant and vanilla DFlash indicates a larger kernel gain of $1.07\times$, consistent with the greater savings at longer verification lengths in Table~\ref{tab:gdnprelim}.

\subsection{GDN kernel microbenchmarks}
\label{sec:gdnresults}


We measure the latency of GDN verification plus accepted-state reconstruction for the 30 GDN layers of Qwen3.6-35B-A3B on one A100, and Table~\ref{tab:gdnprelim} in Appendix~\ref{app:gdnimpl} gives all measured lengths. With every draft token accepted, the \method{} kernels are \mbox{$1.33\times$--$1.35\times$} faster than the default SGLang kernels at lengths 2 and 3 and \mbox{$1.54\times$--$4.04\times$} faster at lengths 4 to 16. The speedup grows with length, consistent with the default kernels writing a full state after every token. With its default launch parameters, as in Table~\ref{tab:main-greedy}, ReplaySSM is slower than the default SGLang kernels at every length, consistent with its lower mean SR than vanilla DFlash. The \method{} kernels are \mbox{$1.95\times$--$6.23\times$} faster than it. These isolated benchmarks start from precomputed projections and exclude the joint producer and two-stage verification. The kernels also save memory, as Appendix~\ref{app:memory} shows. At batch size one, the default verifier reserves 1.88~GB, 40\% of the size of the attention-cache pool, to store the state after every token, whereas the \method{} kernels need no such buffer.

\section{Conclusion}

Choosing the verification length for hybrid GDN--MoE targets trades expected output progress against recurrent and routed-expert work. \method{} makes this choice on the GPU from acceptance estimates and measured target latency and refines it with a lightweight routing correction at the first MoE router. The rest of verification runs only on the selected tokens, with GDN kernels that verify all tokens in parallel and reconstruct every accepted state in one launch. On three GDN--MoE target models in single-request greedy decoding, \method{} is $1.75\times$--$1.88\times$ faster than autoregressive decoding and $1.30\times$--$1.33\times$ faster than vanilla DFlash.

\subsection*{AI Use Statement}

The authors used generative AI tools for manuscript writing and language editing, coding assistance and code review, and checking reported results against experimental records. AI-generated output was not treated as evidence. The authors reviewed all AI-assisted text, code, and checks and take full responsibility for the content of this paper.

\subsection*{Ethics Statement}

This work studies language-model inference efficiency and does not involve human-subject research or the collection of personal data. Cheaper inference lowers the cost of both beneficial and harmful uses of the underlying models. Because \method{} preserves the target's outputs in exact arithmetic, safety, privacy, and fairness remain properties of the target model and its deployment.

\subsection*{Reproducibility Statement}

Sections~\ref{sec:setting}--\ref{sec:evaluation} specify the problem formulation, method, correctness conditions, and reported measurements. Appendices~\ref{app:methoddetails}--\ref{app:e2eprotocol} provide the round procedure, calibration details, derivations and proofs, the profiling and kernel-benchmark setups, and the end-to-end evaluation details. We will release the code upon acceptance.

\newpage
\bibliography{references}
\bibliographystyle{iclr2027_conference}
\clearpage
\appendix
\section{HybridSpec Inference Details}
\label{app:methoddetails}

\subsection{Round procedure}
\label{app:roundprocedure}

Algorithm~\ref{alg:hybridspec} lists one round of Section~\ref{sec:round}, and its comments give the corresponding steps of \mbox{Figure}~\ref{fig:hybridspec-overview}. The lengths $w_0$, $u$, and $w$ stay in GPU memory, and lines~\ref{alg:stage0} and~\ref{alg:stage1} launch the CUDA Graphs for lengths $u$ and $w$ through conditional graph nodes. Under temperature sampling, the point-mass rule of Appendix~\ref{app:sampling} replaces lines~\ref{alg:argmax} and~\ref{alg:accept}, and all other lines are unchanged. At an EOS token or the output limit, the emitted and committed tokens are truncated as in Section~\ref{sec:semantics}.

\begin{algorithm}[htbp]
\caption{One round of \method{}}
\label{alg:hybridspec}
\begin{algorithmic}[1]
\REQUIRE anchor $x_0$, target caches and GDN states, drafter context, and length limit $w_{\mathrm{rem}}$ from the remaining output budget
\item[\textbf{Parameters:}] calibrated acceptance model, $C_{\mathrm{tgt}}$, $\rho$, $s$, $\eta$, and the tables $\bar n_{\mathrm{all}}(w_0)$ and $\bar n_{\mathrm{all}}(w_0\mid n)$
\ENSURE emitted tokens, next anchor, and updated caches, GDN states, and drafter context
\STATE Draft the proposal block $(d_1,\ldots,d_D)$ with the top-1 confidence and entropy of each token
\STATE Estimate $\hat c_1,\ldots,\hat c_D$ and set $\hat p_i\gets\prod_{j=1}^{i}\hat c_j$ for $i=1,\ldots,D$ \COMMENT{App.~\ref{app:costfeatures}}
\STATE $w_0\gets\argmax_{w\in\mathcal W_0}\big[1+\sum_{i=1}^{w-1}\hat p_i-\rho\,C_{\mathrm{tgt}}(w)\big]$ \COMMENT{Eq.~\eqref{eq:nominal-selector}, step 1}
\STATE $u\gets\min(w_0+1,w_{\mathrm{rem}})$ \COMMENT{Stage-0 length}
\STATE Stage~0: run Layer~0 through its MoE router on the first $u$ tokens of $(x_0,d_1,\ldots,d_D)$, retaining compact factors, hidden states, residuals, expert assignments, and routing weights \label{alg:stage0} \COMMENT{steps 2--3}
\STATE $w_{\mathrm c}\gets w_0$ \COMMENT{step 4}
\IF[Stage~0 covers position $w_0+1$]{$u=w_0+1$}
\STATE $n_{\mathrm{new}}\gets U_0(w_0+1)-U_0(w_0)$
\STATE $\Delta L_{\mathrm c}\gets C_{\mathrm{tgt}}(w_0+1)-C_{\mathrm{tgt}}(w_0)-s+\eta\,\big(\bar n_{\mathrm{all}}(w_0\mid n_{\mathrm{new}})-\bar n_{\mathrm{all}}(w_0)\big)$ \COMMENT{Eq.~\eqref{eq:routing-correction}}
\STATE $w_{\mathrm c}\gets w_0+\ind\big[\hat p_{w_0}>\rho\,\Delta L_{\mathrm c}\big]$
\ENDIF
\STATE $w\gets\min(w_{\mathrm c},w_{\mathrm{rem}})$ \COMMENT{executed verification length}
\STATE Stage~1: resume Layer~0 at its experts for the first $w$ tokens, run the remaining layers, and retain their GDN compact factors \label{alg:stage1} \COMMENT{steps 5--6}
\STATE $y_i\gets$ target argmax after $(x_0,d_1,\ldots,d_{i-1})$ for $i=1,\ldots,w$ \label{alg:argmax}
\STATE $A\gets\max\{a\in\{0,\ldots,w-1\}: d_i=y_i \text{ for all } i\leq a\}$ and $x_{\mathrm{new}}\gets y_{A+1}$ \label{alg:accept}
\STATE Advance the attention and convolution caches and the drafter context through the $m=A+1$ committed tokens \COMMENT{step 7}
\STATE Reconstruct the accepted state of every GDN layer in one kernel launch \COMMENT{Eq.~\eqref{eq:commit}, step 8}
\RETURN $(d_1,\ldots,d_A,x_{\mathrm{new}})$, with $x_{\mathrm{new}}$ as the next anchor
\end{algorithmic}
\end{algorithm}

\subsection{GPU-resident selection of verification length}
\label{app:costfeatures}

\paragraph{Draft-side selector.}
The selector reads the top-1 confidence and the entropy $h_i$ of each draft token from GPU memory. It calibrates the confidence per position with isotonic regression. Let $\bar h_i=(h_i+\bar h_{i-1})/2$, with $\bar h_0=h_1$, be an exponential moving average of the entropy within the proposal block. A logistic model then combines the calibrated confidence, the position, their interaction, $\log(1+h_i)$, and the entropy change $\log(h_i/\bar h_{i-1})$. Its outputs are the conditional acceptance probabilities $\hat c_i$. One kernel forms the prefix probabilities and evaluates Equation~\eqref{eq:nominal-selector} for all nominal verification lengths. All selector parameters are fitted on the calibration prompts (Appendix~\ref{app:training}).

\paragraph{Cost calibration.}
Runs on the evaluated runtime fix the verification length at each supported value and measure the mean decode time per round on the calibration prompts. A quadratic in $w$ is fitted to these times, and $C_{\mathrm{tgt}}(w)$ is the fit minus a fixed per-round cost that does not depend on $w$. Calibration also measures $s$, the latency that one more token adds to the Layer-0 GDN layer and router, as the slope of Stage-0 latency in the number of tokens. The constant $\eta$ is the latency that one more distinct expert adds to one MoE layer at batch size one. Every MoE layer of the evaluated models routes each token to 8 of 256 experts. Two measurements support the per-expert model. For Qwen3.6-35B-A3B at verification length 16, the target latency of one verification call in each of 80 requests has a correlation of 0.999 with that call's number of distinct (layer, expert) pairs. For the Qwen3.5-122B-A10B MoE kernel with 16 input rows, and hence 128 token--expert assignments, median latency grows almost linearly with the number of distinct experts, from 0.11~ms at 8 experts to 1.18~ms at 128. Because the number of assignments is fixed, token work adds at most the fitted intercept of 0.06~ms. The fitted slopes give $\eta$, $4.18\,\mu$s for the two 35B models, which share one architecture, and $8.8\,\mu$s for Qwen3.5-122B-A10B. The table value $\bar n_{\mathrm{all}}(w_0)$ is the sum of two averages for the token at position $w_0+1$ on the calibration prompts: the number of new Layer-0 experts and, as Appendix~\ref{app:widthprofile} describes, the number of new (layer, expert) pairs in the later layers. The conditional value $\bar n_{\mathrm{all}}(w_0\mid n)$ replaces the first average with $n$ and the second with its average over the tokens at that position that add $n$ new Layer-0 experts. Cells with fewer than 30 tokens use a linear fit in $n$ for each position. The rate $\rho$ is the fixed point of Dinkelbach's iteration~\citep{dinkelbach1967fractional}. It equals the estimated pooled throughput of the draft-side selector on the rounds of the capture run that Appendix~\ref{app:training} describes. Its numerator sums $1+\sum_{i=1}^{w_0-1}\hat p_i$ over the rounds, and its denominator sums $C_{\mathrm{tgt}}(w_0)$ plus the fixed per-round cost removed from $C_{\mathrm{tgt}}$.

\section{Analysis of the Throughput Objective}
\label{app:analysis}
\label{app:width}

Because $A_D$ is an integer, $\min(A_D,M)=\sum_{i=1}^M\ind[A_D\geq i]$. Taking a conditional expectation and adding the token emitted after the accepted prefix gives $G$ for nonterminal rounds (Section~\ref{sec:semantics}).

For $M'>M$ at a common state, suppress $\mathcal I$ and let $\Delta G=G(M')-G(M)=\sum_{i=M+1}^{M'}p_i$ and $\Delta L=L(M')-L(M)$. With positive latencies, $M'$ has a higher $G/L$ exactly when
\begin{equation}
 \Delta G\,L(M)>G(M)\,\Delta L.
 \label{eq:marginal}
\end{equation}
Replacing the current rate $G(M)/L(M)$ with the pooled rate $\rho$ turns Equation~\eqref{eq:marginal} into the additive test $\Delta G>\rho\,\Delta L$. This is the form of both the draft-side score in Equation~\eqref{eq:nominal-selector} and the routing correction in Equation~\eqref{eq:routing-correction}. Under the additive test, even a draft token that is certain to be accepted helps only if its added latency is below $1/\rho$, the pooled latency per emitted token. Because the measured latency curve can be nonconvex or even nonmonotone, the selector scores every nominal verification length rather than stopping at the first unfavorable comparison.

For serial rounds at $B=1$, the throughput of a stationary policy $\pi$ converges, under standard ergodicity conditions, to $\E_\pi[R]/\E_\pi[T_{\mathrm{round}}]$, where $T_{\mathrm{round}}$ is the round latency. This limit is a ratio of expectations rather than an average of per-round ratios, so Section~\ref{sec:model} defines the objective as pooled throughput. During decoding, later rounds depend on earlier choices, so $\rho$ only approximates the optimal pooled throughput of the decoding process.

\section{Correctness Analysis}
\label{app:correctness}

\subsection{Triangular GDN factorization and state reconstruction}
\label{app:factors}

By induction on Equation~\eqref{eq:gdn}, Equation~\eqref{eq:unrolled} holds for every $i$. Substituting it for $S_{i-1}$ into $z_i$ and using $\alpha_ie^{g_{i-1}}=e^{g_i}$ gives
\begin{equation}
 z_i+\sum_{j<i}\beta_i e^{g_i-g_j}(k_i^\top k_j)z_j
 =\beta_i v_i-\beta_i e^{g_i}S_0k_i.
 \label{eq:triangularrow}
\end{equation}
With the matrices of Section~\ref{sec:gdnbackground}, stacking Equation~\eqref{eq:triangularrow} over $i=1,\ldots,w$ gives the triangular system in Equation~\eqref{eq:blockoutputs}. In matrix form, $H=I+\operatorname{tril}_{-1}\big(\operatorname{diag}(\beta)(\Gamma\odot KK^\top)\big)$. Here $\Gamma$ is the decay-aware causal mask of Gated DeltaNet~\citep{yang2025gated}, with $\Gamma_{ij}=e^{g_i-g_j}$ for $j\leq i$ and $0$ otherwise, $\odot$ is the elementwise product, and $\operatorname{tril}_{-1}$ keeps the entries below the diagonal.

The first $m$ equations depend only on the leading $m\times m$ block of $H$ and the first $m$ rows of the right-hand side. Thus appending or removing later tokens cannot change the compact factors of the prefix in exact arithmetic. Equation~\eqref{eq:unrolled} at $i=m$ is Equation~\eqref{eq:commit} written as a sum, so reusing the leading factors reconstructs the correct accepted state.

Substituting Equation~\eqref{eq:unrolled} into $o_i=S_i q_i$ gives $o_i=e^{g_i}S_0q_i+\sum_{j=1}^{i}e^{g_i-g_j}z_j(k_j^\top q_i)$. Stacking $q_i^\top$ and $o_i^\top$ as the rows of $Q\in\R^{w\times d_k}$ and $O\in\R^{w\times d_v}$ gives
\begin{equation}
 O=\operatorname{diag}(e^{g})QS_0^\top+(\Gamma\odot QK^\top)Z.
 \label{eq:blockout}
\end{equation}
Once Equation~\eqref{eq:blockoutputs} yields $Z$, Equation~\eqref{eq:blockout} evaluates all GDN outputs $o_1,\ldots,o_w$ with dense matrix products. These outputs match the recurrent computation. A numerical implementation still requires stable accumulation and appropriate handling of extreme decay.

\subsection{Correctness of prefix truncation under greedy decoding}
\label{app:greedy}

The analysis uses five assumptions. Arithmetic is exact. Causal prefix invariance holds, meaning that removing suffix tokens does not change any target operator's computation for earlier tokens. Prefix states are recovered exactly: after each round, the commit sets every GDN recurrent state, convolution cache, and attention cache to the target's state after the committed tokens. Processing several tokens in one forward pass gives the same per-token results as processing them one at a time. Stage~1 resumes Layer~0 from the retained Stage-0 values of the first $w$ tokens and reads no retained value of a later token. The fourth assumption holds for attention with a key--value cache, for the causal convolution of GDN, for position-wise layers, and for top-$k$ MoE routing and expert computation without capacity limits. It also holds for the GDN recurrence, because Appendix~\ref{app:factors} shows that computing the updates of a block of tokens at once gives the same outputs and states as the token-by-token recurrence. These assumptions exclude MoE capacity rules that change an earlier token's expert computation when later tokens are removed.

\begin{proposition}
\label{prop:greedy}
Under these assumptions, greedy decoding with \method{} emits the same tokens as target-only greedy decoding for any verification length that each round selects, provided that both stop at the first EOS token or at the output limit.
\end{proposition}

\begin{proof}
The proof maintains an invariant. After prefill and at the start of each round, the emitted tokens form a prefix of the target-only greedy continuation computed without stopping, and every persistent cache holds the state that the target alone computes for the tokens before the anchor. By causal prefix invariance and because Stage~1 reads only the retained values of the first $w$ tokens, the extra Stage-0 token does not change the target logits for the anchor and the retained draft tokens. Because $m\leq w$, its retained factors never enter the commit. Because every cache holds this state and a multi-token pass matches one-token passes, the logits at the anchor equal those of target-only decoding. If the first draft token matches their argmax, accepting it extends the prefix, and the same argument applies to each later accepted draft token. At the first mismatch, or after the last retained draft token, the target output at that position is the next token of target-only decoding. Because prefix states are recovered exactly, committing exactly the anchor and the accepted draft tokens restores the invariant for the next round. Each round emits at least one token, so both processes truncate the same continuation at the first EOS token or at the output limit.
\end{proof}

In floating point, the multi-token kernels compute the same quantities with a different operation order and precision. This can flip the argmax when logits are nearly tied, after which later tokens can also differ. Appendix~\ref{app:sampling} reuses the cache part of the invariant for temperature sampling.

\subsection{Exactness under temperature sampling}
\label{app:sampling}

Fix the prompt and consider a round with anchor $x_0$ and proposal prefix $d_1,\ldots,d_M$. For ${i=1,\ldots,M+1}$, let $p^{\mathrm t}_i$ be the sampling distribution that the verifier computes at position $i$, that is, for the next token after the prompt, the tokens emitted before the round, and $d_1,\ldots,d_{i-1}$. It includes temperature scaling and any top-$k$ or top-$p$ filtering.

The verifier applies speculative sampling~\citep{leviathan2023fast,chen2023sampling} with each draft token treated as a proposal of probability one. We call this the point-mass rule. For $i=1,\ldots,M$, the verifier draws a uniform random number $\xi_i\in[0,1]$ and accepts $d_i$ if $\xi_i\leq p^{\mathrm t}_i(d_i)$. At the first rejection, it emits a token drawn from $p^{\mathrm t}_i$ with $d_i$ removed and the remaining probabilities renormalized. The round then ends. If all $M$ draft tokens are accepted, it emits a token drawn from $p^{\mathrm t}_{M+1}$. We assume that each draw is independent of everything determined before it. SGLang implements this rule with its default acceptance thresholds, which our configuration uses.

Let $y_i$ be the token that the round emits at position $i$. Let $\mathcal F_i$ be everything determined before the draws at position $i$, including the proposal block, $M$, and all earlier tokens and draws. If the round reaches position $i$, that is, if $i\leq M+1$ and $d_1,\ldots,d_{i-1}$ are accepted, then
\begin{equation}
 \Prb(y_i=v\mid\mathcal F_i)=p^{\mathrm t}_i(v)\quad\text{for every token } v.
 \label{eq:pointmass}
\end{equation}
The draws at position $i$ are independent of $\mathcal F_i$. If $i=M+1$, the verifier draws $y_i$ from $p^{\mathrm t}_{M+1}$. If $i\leq M$, the token $d_i$ is emitted only through acceptance, which has probability $p^{\mathrm t}_i(d_i)$. Any other token $v$ requires a rejection, with probability $1-p^{\mathrm t}_i(d_i)$, and then a draw of $v$, with probability $p^{\mathrm t}_i(v)/\big(1-p^{\mathrm t}_i(d_i)\big)$. The product is $p^{\mathrm t}_i(v)$. If $p^{\mathrm t}_i(d_i)=1$, a rejection never occurs and $p^{\mathrm t}_i(v)=0$ for every $v\neq d_i$, so the equation still holds. Because Equation~\eqref{eq:pointmass} holds for every proposal block and every $M$, the verification length changes only how a token is produced, not its distribution.

\begin{proposition}
\label{prop:sampling}
Adopt the assumptions of Proposition~\ref{prop:greedy}. Assume also that verification uses the point-mass rule and that each round fixes its proposal block and $M$ before its verification draws. Both may depend on the prompt and on everything determined in earlier rounds. Assume further that generation stops at the first EOS token or at the output limit, as in target-only sampling. Then the output has the same distribution as target-only sampling.
\end{proposition}

\begin{proof}
The cache part of the invariant in Appendix~\ref{app:greedy} does not use the acceptance rule. At the start of each round, every cache therefore holds the state that the target alone computes for the tokens before the anchor. With the other assumptions, and because the verifier applies the same temperature and filtering as target-only sampling, $p^{\mathrm t}_i$ is the target-only sampling distribution given the prompt and all tokens emitted before position $i$. By Equation~\eqref{eq:pointmass}, each emitted token follows this distribution given $\mathcal F_i$. This distribution depends only on the earlier emitted tokens, which $\mathcal F_i$ contains, so the same holds given only these tokens. The round and position that produce each emitted token are determined before its draws, so this holds for every emitted token in turn. By the chain rule, every sequence of emitted tokens has the same probability as under target-only sampling. Both processes apply the same stopping rule, so their outputs have the same distribution.
\end{proof}

Like Proposition~\ref{prop:greedy}, this result assumes exact arithmetic. In floating point, $p^{\mathrm t}_i$ differs slightly from the target-only sampling distribution, so the two output distributions agree only approximately.

In \method{}, the draft tokens are the drafter's top-1 tokens, and the verification length depends on the drafter's confidence and entropy values, the Layer-0 expert assignments of the Stage-0 tokens, and the remaining output budget, but on no verification draw. Proposition~\ref{prop:sampling} therefore covers the full length policy, including the routing correction.

The argument depends on the point-mass rule. The general rule~\citep{leviathan2023fast,chen2023sampling} samples $d_i$ from a draft distribution $q^{\mathrm d}_i$, accepts it with probability $\min\{1,p^{\mathrm t}_i(d_i)/q^{\mathrm d}_i(d_i)\}$, and otherwise draws from the normalized positive part of $p^{\mathrm t}_i-q^{\mathrm d}_i$. Under that rule, the emitted token follows $p^{\mathrm t}_i$ only on average over $d_i\sim q^{\mathrm d}_i$, not for each value of $d_i$. A length rule that reads $d_i$ can then bias the output, because the draft tokens that reach verification may no longer follow $q^{\mathrm d}_i$. The routing correction is such a rule, because it reads the expert assignment of $d_{w_0}$ before deciding whether to add that token. The point-mass rule removes this dependence, since Equation~\eqref{eq:pointmass} holds for every draft token.

\section{Experimental Details}
\label{app:experimental}

\subsection{Profile across verification lengths}
\label{app:widthprofile}

The profile used Qwen3.6-35B-A3B with its DFlash drafter, SGLang 0.5.16, and one NVIDIA A100 GPU, at batch size one with a context limit of 8,192 tokens. SGLang used its default Triton~\citep{tillet2019triton} MoE configuration, because no tuned configuration was available for this model. The workload was ten GSM8K test prompts~\citep{cobbe2021gsm8k}, each decoded greedily for 32 new tokens. Each run set the DFlash block size to the verification length $w$, so $D=M=w-1$.

At $w=4$, 8, and 16, mean profiled GPU time per measured verification call was 1.795, 1.817, and 1.856~ms for the drafter and 9.527, 12.648, and 16.318~ms for target verification. The target-to-drafter time ratios were therefore $5.3\times$, $7.0\times$, and $8.8\times$. From $w=4$ to 16, drafter time increased by 3.4\%, whereas target time increased by 71.3\%. Drafter and target intervals may overlap under the scheduler, so these values do not form an end-to-end latency decomposition.

\begin{table}[htbp]
\centering\scriptsize
\caption{Qwen3.6-35B-A3B target-side GPU time across verification lengths, in milliseconds per measured verification call. The total is the union of slightly overlapping component intervals, so the components do not sum exactly to it.}
\label{tab:widthfull}
\begin{tabular}{@{}rrrrrrr@{}}
\toprule
$w$ & \shortstack{Measured\\calls} & \shortstack{Target\\total} & GDN & MoE & \shortstack{Softmax\\attention} & Other \\
\midrule
4  & 101 & 9.527  & 2.498 & 5.331  & 0.686 & 1.019 \\
5  & 85  & 10.055 & 2.659 & 5.691  & 0.696 & 1.017 \\
6  & 80  & 10.811 & 2.835 & 6.248  & 0.710 & 1.026 \\
7  & 74  & 11.876 & 2.984 & 7.171  & 0.705 & 1.026 \\
8  & 72  & 12.648 & 3.096 & 7.826  & 0.698 & 1.037 \\
10 & 71  & 14.075 & 3.413 & 8.934  & 0.705 & 1.033 \\
12 & 71  & 14.659 & 3.734 & 9.165  & 0.719 & 1.050 \\
15 & 71  & 16.078 & 4.224 & 10.085 & 0.722 & 1.057 \\
16 & 70  & 16.318 & 4.350 & 10.187 & 0.725 & 1.065 \\
\bottomrule
\end{tabular}
\end{table}

Across the nine measured verification lengths, an unweighted ordinary least-squares fit to GDN time gives $1.89+0.154w$~ms, with coefficient of determination $R^2=0.999$. A log--log fit to MoE time gives $2.66w^{0.50}$~ms (log-space $R^2=0.976$). From $w=4$ to 16, profiled GPU time for GDN and MoE increased by 74.1\% and 91.1\%, whereas softmax-attention and other target-side time increased by 5.7\% and 4.5\%. Each verification length was measured in a separate run, so the fitted trends do not isolate the effect of length at a fixed decoding state.

\paragraph{Dense-target comparison.}
A separate comparison on one NVIDIA A100 GPU paired a dense Qwen3-4B target with DFlash at context lengths of 2,048, 8,192, and 32,768 tokens. Over 15 samples at each setting, the median latency of its captured target-verification CUDA Graph was 3.98\%, 5.44\%, and 6.08\% higher at $w=16$ than at $w=4$ for the three context lengths.

\paragraph{Layer-0 routing as a proxy.}
To estimate the later-layer counts, a separate run on the calibration prompts verifies at length 16 with greedy decoding and at most 1,024 new tokens and records the top-8 experts of every token in every layer. For each draft position, let $n_0$ be the number of new Layer-0 experts that the token adds and $n_{>0}$ the number of new (layer, expert) pairs that it adds in layers 1--39. After the mean of each position is subtracted, $n_0$ and $n_{>0}$ have a correlation of 0.49 on Qwen3.6-35B-A3B and 0.52 on Qwen3.5-35B-A3B, and a least-squares fit gives 9.0 and 10.0 later-layer pairs per new Layer-0 expert.

\subsection{Kernel microbenchmark details}
\label{app:gdnimpl}

The benchmark runs all 30 GDN layers of Qwen3.6-35B-A3B at batch size one, starting from random bfloat16 (BF16) projections and FP32 states. It captures the verification and the state commit of each implementation as CUDA Graphs. Instead of modeling acceptance, it fixes the number of committed tokens at ${m=w}$, where every draft token is accepted, or at ${m=\lceil w/2\rceil}$. Each latency is the median of 30 samples of 10 graph replays. ReplaySSM uses its SGLang kernels with their default launch parameters, as in the end-to-end evaluation, and a ring buffer of at least twice the verification length. Because ReplaySSM folds this buffer into the state only when the buffer is flushed, its verification latency is the mean over 300 alternating verification and commit rounds. At the lengths in Table~\ref{tab:gdnprelim}, a second sweep reproduced every total latency within 0.5\%, except the default kernels at length~2, which differed by 13\%. ReplaySSM was slower than the default SGLang kernels at every length. In a separate run with launch parameters tuned for batch size one, ReplaySSM was \mbox{$1.05\times$--$1.71\times$} faster than the default kernels over lengths 2 to 16, and \method{} remained \mbox{$1.16\times$--$2.42\times$} faster than it. At ${m=\lceil w/2\rceil}$, the latency of the default kernels changes by less than 0.2\%, because they write a state after every token during verification. The latencies of \method{} and ReplaySSM fall by \mbox{2\%--6\%} and \mbox{14\%--31\%}, respectively, because both fold only the committed tokens into the state. The all-accepted speedups over the default kernels in Section~\ref{sec:gdnresults} are therefore conservative, whereas the all-accepted setting favors \method{} against ReplaySSM.

\begin{table}[htbp]
\centering\scriptsize
\caption{Latency (ms) of GDN verification plus accepted-state reconstruction for the 30 GDN layers of Qwen3.6-35B-A3B on one A100 when all $m=w$ input tokens are committed, that is, when every draft token is accepted. Speedups of \method{} are given for $m=w$ and for $m=\lceil w/2\rceil$, when about half of the input tokens are committed. ReplaySSM uses its default launch parameters, as in Table~\ref{tab:main-greedy}.}
\label{tab:gdnprelim}
\begin{tabularx}{\linewidth}{@{}cCCCCC@{}}
\toprule
 & \multicolumn{3}{c}{Latency (ms)} & \multicolumn{2}{c}{Speedup ($\times$), $m=w$ / $m=\lceil w/2\rceil$} \\
\cmidrule(lr){2-4}\cmidrule(l){5-6}
$w$ & SGLang & ReplaySSM & \method{} & vs.\ SGLang & vs.\ ReplaySSM \\
\midrule
2  & 0.4726 & 0.6926 & 0.3552 & $1.33$ / $1.35$ & $1.95$ / $1.66$ \\
3  & 0.5012 & 0.8058 & 0.3721 & $1.35$ / $1.38$ & $2.17$ / $1.90$ \\
4  & 0.5920 & 0.9192 & 0.3847 & $1.54$ / $1.62$ & $2.39$ / $1.90$ \\
8  & 0.9733 & 1.5050 & 0.4366 & $2.23$ / $2.37$ & $3.45$ / $2.55$ \\
12 & 1.4263 & 2.4859 & 0.4491 & $3.18$ / $3.38$ & $5.54$ / $4.59$ \\
16 & 1.9111 & 2.9451 & 0.4728 & $4.04$ / $4.29$ & $6.23$ / $4.56$ \\
\bottomrule
\end{tabularx}
\end{table}

\paragraph{Triangular solve for short inputs.}
The matrix $H$ in Equation~\eqref{eq:blockoutputs} is unit lower triangular with at most 16 rows, so $X=I-H$ is strictly lower triangular and $X^{16}=0$. Lengths 4, 6, and 8 use closed-form inverses of $4\times4$ blocks, and lengths 6 and 8 join two such blocks with one correction term. Lengths 9 to 16 use the exact identity $H^{-1}=(I+X+X^2+X^3)(I+X^4+X^8+X^{12})$, which needs only matrix products. The other lengths, 2, 3, 5, and 7, form the inverse row by row in at most six steps on a matrix with at most seven rows, and each step computes a whole row at once. For every length, all heads run in parallel, and the products with keys, values, and $S_0$ are matrix products over all tokens of the block, so verification remains block-parallel.

\paragraph{Kernel correctness.}
Before timing, the benchmark compares the \method{} kernels with the default SGLang kernels on one GDN layer at lengths 2, 4, 8, and 16. Each comparison runs two consecutive rounds that commit $w$ and $w/2$ tokens, in both orders. The largest relative error in Frobenius norm is $4.3\times10^{-3}$ for outputs and $1.7\times10^{-3}$ for the reconstructed state. With FP32 instead of BF16 arithmetic in the \method{} kernels, these errors fall below $10^{-4}$ and $10^{-6}$.

\subsection{Correctness and task quality}
\label{sec:correctnessresults}

We check correctness in two steps, both on Qwen3.6-35B-A3B under greedy decoding. The first step tests whether splitting verification at the router boundary changes outputs. It compares \method{} with and without the split, where the unsplit form runs all of Layer~0 before choosing the verification length. On 16 prompts that each generate 1,024 tokens, \method{} runs 3,114 target forward passes with verification lengths 3 to 16 and produces the same output tokens, acceptance histograms, and numbers of rounds in both forms. It never reads the selected length on the CPU. The second step tests task quality. On all 1,319 GSM8K test problems~\citep{cobbe2021gsm8k}, \method{} solves 96.06\% and vanilla DFlash 96.21\%. The one-sided 95\% lower confidence bound of the paired difference, $-0.71$ points, is above the noninferiority margin of $-1$ point fixed before evaluation. The two systems do not emit identical tokens, which reordered floating-point kernels can cause by flipping near-tied logits, so this step compares task quality rather than exact token agreement.

\subsection{Memory for verification states}
\label{app:memory}

Because the default SGLang verifier stores the GDN state after every token, it preallocates, for each request slot, one state per GDN layer for every position of the full target input. For Qwen3.6-35B-A3B, whose 30 GDN layers each hold a $32\times128\times128$ state in FP32, this buffer takes $16\times30\times32\times128\times128\times4$~bytes, or 0.94~GB, per slot. At batch size one, SGLang allocates two slots, and the server log of our runs on one A100 reports 1.88~GB for the buffer. The same log reports a pool of 112,687 tokens for the attention caches of the target and the drafter, which take about 44~KB per token. The buffer therefore equals the caches of about 45,000 tokens, 40\% of the pool. The \method{} kernels retain only the compact factors, which take about 0.01~GB for 16 tokens across the 30 layers even in FP32, so they do not need this buffer. Releasing it raises the maximum context length that fits in memory by about 40\%.

The saving grows with the number of concurrent requests, because SGLang allocates one slot per request plus one. With four concurrent requests, vanilla DFlash allocates 4.69~GB for the buffer and has room for 12,472 cached tokens, whereas the batched version of \method{} in Appendix~\ref{app:batching} allocates no buffer and caches 122,512 tokens.

\section{End-to-End Evaluation Details}
\label{app:e2eprotocol}

\subsection{Models, baselines, and metrics}

The targets are the official Qwen checkpoints of Qwen3.5-35B-A3B and Qwen3.6-35B-A3B in BF16 and of Qwen3.5-122B-A10B in \mbox{GPTQ-Int4}. The drafters are the official DFlash checkpoints that z-lab released for these targets, and all checkpoints come from Hugging Face. Each pair is calibrated and evaluated separately. Qwen3.5-122B-A10B uses two-way tensor parallelism across two GPUs without NVLink. The prompts come from seven of the eleven SPEED-Bench Qualitative categories (Section~\ref{sec:setup}), and a seeded random split divides their 560 prompts into the calibration and test halves. Each request generates at most 1,024 new tokens, as in the SPEED-Bench evaluation of \citet{fu2026nemotronlabsdiffusion}, and thinking mode is enabled.

\paragraph{Baselines.}
\mbox{D-Cut} and EVICT follow their papers. At batch size one, the budget allocation of \mbox{D-Cut} reduces to a per-request pruning depth. As in its paper, \mbox{D-Cut} restricts this budget to four fixed fractions of the full target input, so it chooses among verification lengths 4, 8, 12, and 16. The draft tree of EVICT reduces to the single DFlash chain. EVICT + ReplaySSM applies the truncation of EVICT and verifies the retained prefix with ReplaySSM. The drafter generates all $D$ draft tokens before any length decision, so all speculative systems incur the same drafting cost.

\paragraph{Metrics.}
Pooled decode throughput (TPS) is the number of tokens emitted after the prefill-produced anchor divided by the wall-clock decode time, pooled over the requests of a category. The decode time includes CPU-side control, graph dispatch, synchronization, and padding. SR is the ratio of pooled TPS to that of AR on the same prompts. AL is the number of emitted tokens divided by the number of verification rounds, which corresponds to $G$ in Section~\ref{sec:semantics}. Category averages are arithmetic means over the seven categories.

\subsection{Comparisons and ablations}

\paragraph{End-to-end comparisons.}
All systems share prompts, stopping rules, precision, and output limits. Systems that select a verification length use the calibration split and a cost function calibrated for their own runtime. Each system runs in three sessions, each with a fresh server process, and the sessions are pooled.

\paragraph{Ablations.}
Each ablation keeps the drafter, prompts, precision, and calibration split of \method{}. To select the calibrated static length, a sweep on a subset of the calibration prompts keeps up to three fixed lengths within 5\% of the fastest. These candidates and length 16 then run on all calibration prompts, and the one with the highest pooled throughput is used. The GDN variant replaces the \method{} verification and commit kernels with the default SGLang kernels.

\subsection{Calibration data}
\label{app:training}

\paragraph{Calibration runs.}
Calibration runs separately for each target--drafter pair and uses only the calibration prompts, which are disjoint from the test prompts. A capture run verifies all $D$ draft tokens in every round with greedy decoding. It records the drafter's confidences and entropies, the number of new Layer-0 experts that each draft token adds, and the number of accepted draft tokens. These rounds train the acceptance model, give the Layer-0 averages in $\bar n_{\mathrm{all}}$, and are the rounds over which $\rho$ is computed. Runs with a fixed verification length measure $C_{\mathrm{tgt}}$, and a separate run at length 16 gives the later-layer averages, as Appendices~\ref{app:costfeatures} and~\ref{app:widthprofile} describe.

\paragraph{Acceptance-model training.}
Because the capture run verifies all $D$ draft tokens, a shorter verification never censors the labels of later draft tokens. Training $\hat c_i$ uses the label $\ind[A_D\geq i]$ among rounds with $A_D\geq i-1$. Rounds cut short by EOS or the output limit are excluded.

\subsection{Correctness checks}
\label{app:quality}

The GSM8K check compares \method{} with vanilla DFlash. \method{} solves 1,267 problems and vanilla DFlash solves 1,269, a paired difference of $-0.15$ percentage points. The lower confidence bound is a normal approximation from the 20 problems on which the two systems disagree, 11 solved only by vanilla DFlash and 9 only by \method{}. A second paired run gives 96.21\% without the routing correction and 96.13\% with it.

The router-boundary check uses 16 prompts spanning math, coding, writing, and chat, and each request ignores EOS and generates exactly 1,024 tokens. The split and unsplit forms also process the same numbers of tokens before and after the length decision. After Stage~0 finishes, overwriting its retained tensors for tokens beyond the selected length leaves every output unchanged. The run records no graph recapture and no rerun of the Layer-0 router.

\subsection{Unseen categories}
\label{app:ood}

The four SPEED-Bench Qualitative categories outside the study, humanities, QA, STEM, and multilingual, contribute all 80 of their prompts each. The prompts are rendered as in the main evaluation, and none matches a study prompt. \method{} uses the parameters of Table~\ref{tab:main-greedy} without recalibration. \method{} is faster than vanilla DFlash in every category, by $1.28\times$ for multilingual to $1.31\times$ for QA. Its mean verification length is 5.7, compared with 6.2 on the test prompts.

\subsection{Temperature-one sampling}
\label{app:sampling-results}

Table~\ref{tab:main-sampling} repeats the comparison of Table~\ref{tab:main-greedy} for the two 35B pairs under temperature-one sampling without top-$p$ or top-$k$ truncation. \method{} keeps the parameters calibrated under greedy decoding. Its mean SR is $1.57\times$ on Qwen3.5-35B-A3B and $1.60\times$ on Qwen3.6-35B-A3B, compared with $1.20\times$ and $1.21\times$ for vanilla DFlash and $1.34\times$ and $1.38\times$ for EVICT, the strongest baseline. \method{} has the highest SR in every category. Sampling lowers the AL of every speculative method, so every SR is lower than under greedy decoding. The lead of \method{} over vanilla DFlash is $1.31\times$ and $1.32\times$, compared with $1.30\times$ and $1.33\times$ under greedy decoding.

\begin{table}[H]
\centering\scriptsize
\setlength{\tabcolsep}{1.9pt}
\caption{End-to-end comparison under temperature-one sampling, laid out as in Table~\ref{tab:main-greedy}.}
\label{tab:main-sampling}
\begin{tabular}{@{}lcccccccccccccccc@{}}
\toprule
 & \multicolumn{2}{c}{Coding} & \multicolumn{2}{c}{Math} & \multicolumn{2}{c}{Reasoning} & \multicolumn{2}{c}{RAG} & \multicolumn{2}{c}{Summ.} & \multicolumn{2}{c}{Writing} & \multicolumn{2}{c}{Roleplay} & \multicolumn{2}{c}{Avg.} \\
\cmidrule(lr){2-3}\cmidrule(lr){4-5}\cmidrule(lr){6-7}\cmidrule(lr){8-9}\cmidrule(lr){10-11}\cmidrule(lr){12-13}\cmidrule(lr){14-15}\cmidrule(lr){16-17}
Method & AL & SR & AL & SR & AL & SR & AL & SR & AL & SR & AL & SR & AL & SR & AL & SR \\
\midrule
\multicolumn{17}{@{}l}{\textit{Qwen3.5-35B-A3B}} \\
AR target            & \notapplicable & $1.00\times$ & \notapplicable & $1.00\times$ & \notapplicable & $1.00\times$ & \notapplicable & $1.00\times$ & \notapplicable & $1.00\times$ & \notapplicable & $1.00\times$ & \notapplicable & $1.00\times$ & \notapplicable & $1.00\times$ \\
Vanilla DFlash       & 5.29 & $1.56\times$ & 4.18 & $1.23\times$ & 3.88 & $1.15\times$ & 5.20 & $1.50\times$ & 4.11 & $1.19\times$ & 3.23 & $0.96\times$ & 2.63 & $0.81\times$ & 4.07 & $1.20\times$ \\
\mbox{D-Cut}                & 4.70 & $1.58\times$ & 3.84 & $1.33\times$ & 3.78 & $1.32\times$ & 4.97 & $1.62\times$ & 3.88 & $1.31\times$ & 3.13 & $1.11\times$ & 2.54 & $0.94\times$ & 3.84 & $1.31\times$ \\
EVICT                & 4.81 & $1.61\times$ & 3.97 & $1.37\times$ & 3.86 & $1.35\times$ & 5.03 & $1.63\times$ & 3.96 & $1.33\times$ & 3.17 & $1.13\times$ & 2.56 & $0.96\times$ & 3.91 & $1.34\times$ \\
ReplaySSM            & 5.06 & $1.45\times$ & 3.96 & $1.14\times$ & 4.01 & $1.16\times$ & 5.28 & $1.48\times$ & 4.12 & $1.16\times$ & 3.26 & $0.95\times$ & 2.67 & $0.81\times$ & 4.05 & $1.16\times$ \\
EVICT + ReplaySSM    & 4.75 & $1.57\times$ & 3.86 & $1.34\times$ & 3.81 & $1.32\times$ & 4.89 & $1.58\times$ & 3.96 & $1.33\times$ & 3.12 & $1.10\times$ & 2.54 & $0.94\times$ & 3.85 & $1.31\times$ \\
\oursrow \method{}            & 4.88 & $1.91\times$ & 3.86 & $1.59\times$ & 3.82 & $1.57\times$ & 4.79 & $1.84\times$ & 3.98 & $1.57\times$ & 3.16 & $1.33\times$ & 2.59 & $1.16\times$ & 3.87 & $1.57\times$ \\
\midrule
\multicolumn{17}{@{}l}{\textit{Qwen3.6-35B-A3B}} \\
AR target            & \notapplicable & $1.00\times$ & \notapplicable & $1.00\times$ & \notapplicable & $1.00\times$ & \notapplicable & $1.00\times$ & \notapplicable & $1.00\times$ & \notapplicable & $1.00\times$ & \notapplicable & $1.00\times$ & \notapplicable & $1.00\times$ \\
Vanilla DFlash       & 4.77 & $1.42\times$ & 4.40 & $1.32\times$ & 4.22 & $1.26\times$ & 4.99 & $1.44\times$ & 4.25 & $1.24\times$ & 3.17 & $0.96\times$ & 2.70 & $0.83\times$ & 4.07 & $1.21\times$ \\
\mbox{D-Cut}                & 4.63 & $1.58\times$ & 4.31 & $1.50\times$ & 3.96 & $1.38\times$ & 4.90 & $1.60\times$ & 4.13 & $1.40\times$ & 3.10 & $1.11\times$ & 2.61 & $0.97\times$ & 3.95 & $1.36\times$ \\
EVICT                & 4.73 & $1.61\times$ & 4.25 & $1.49\times$ & 4.06 & $1.43\times$ & 4.86 & $1.60\times$ & 4.12 & $1.40\times$ & 3.09 & $1.12\times$ & 2.62 & $0.99\times$ & 3.96 & $1.38\times$ \\
ReplaySSM            & 4.70 & $1.36\times$ & 4.43 & $1.29\times$ & 4.08 & $1.19\times$ & 4.99 & $1.40\times$ & 4.36 & $1.23\times$ & 3.15 & $0.92\times$ & 2.69 & $0.81\times$ & 4.06 & $1.17\times$ \\
EVICT + ReplaySSM    & 4.65 & $1.53\times$ & 4.26 & $1.44\times$ & 3.99 & $1.36\times$ & 4.87 & $1.55\times$ & 4.14 & $1.36\times$ & 3.07 & $1.07\times$ & 2.62 & $0.95\times$ & 3.94 & $1.32\times$ \\
\oursrow \method{}            & 4.56 & $1.83\times$ & 4.19 & $1.73\times$ & 4.20 & $1.72\times$ & 4.78 & $1.83\times$ & 4.07 & $1.61\times$ & 3.15 & $1.35\times$ & 2.49 & $1.12\times$ & 3.92 & $1.60\times$ \\
\bottomrule
\end{tabular}
\end{table}

\subsection{Batched decoding}
\label{app:batching}

The batched version of \method{} extends the length policy to a batch of $B$ requests. All requests in a round share one verification length, so that the batch runs as one CUDA Graph of fixed shape. The policy chooses this length from $\{2,4,8,12,16\}$ by maximizing $\sum_{b=1}^{B}\hat G_b(w)-\rho_B\,C_B(w)$, where $\hat G_b(w)$ is the expected number of tokens that request $b$ emits at length $w$, $C_B(w)$ is the measured latency of a round with $B$ requests, and $\rho_B$ is the Dinkelbach rate for batches of $B$ requests. At $B=1$, this rule reduces to Equation~\eqref{eq:nominal-selector}. A request that would prefer a shorter length still verifies all $w$ tokens, because the batch already pays for $w$ positions and extra verified tokens can only add accepted tokens. The length is fixed before verification, so the batch preserves the target's outputs in exact arithmetic. Stage~0 and the routing correction carry over: the correction counts the Layer-0 experts that the next candidate length adds to the union over the batch and expands to that length if the added tokens are worth their latency. The latencies $C_B(w)$, the rates $\rho_B$, and the correction parameters are calibrated on the calibration prompts for batches of two and four requests, and a batch of three runs as a padded batch of four.

On the 280 test prompts of Qwen3.6-35B-A3B under greedy decoding, four concurrent clients each send a new request as soon as the previous one finishes. The batched version then emits 497 tokens per second in total, $1.10\times$ the 451 of vanilla DFlash and $1.39\times$ the 358 of autoregressive decoding, while vanilla DFlash with the \method{} GDN kernels emits 472. As at batch size one, it accepts fewer tokens per round than vanilla DFlash, 4.25 versus 4.57, yet emits more tokens per second. With the \method{} kernels, the batched runtime also has room for 122,512 cached tokens, 9.8 times the 12,472 tokens of vanilla DFlash, as Appendix~\ref{app:memory} details. The gain is smaller than at batch size one mainly because the length policy helps less: one length from five candidates must serve all requests, so it follows each request less closely.

\end{document}

%% file: macros.tex
\newcommand{\method}{\mbox{HybridSpec}}
\definecolor{oursrowbg}{HTML}{E5F1E4}
\newcommand{\oursrow}{\rowcolor{oursrowbg}}
\newcommand{\E}{\mathbb{E}}
\newcommand{\Prb}{\mathbb{P}}
\newcommand{\R}{\mathbb{R}}
\newcommand{\ind}{\mathbf{1}}
\DeclareMathOperator*{\argmax}{arg\,max}

\newcommand{\notapplicable}{\textit{n/a}}

\newtheorem{proposition}{Proposition}
\newcolumntype{Y}{>{\raggedright\arraybackslash}X}
\newcolumntype{C}{>{\centering\arraybackslash}X}
\floatstyle{ruled}
\newfloat{algorithm}{htbp}{loa}
\floatname{algorithm}{Algorithm}

%% file: references.bib
@inproceedings{leviathan2023fast,
 title={Fast Inference from Transformers via Speculative Decoding},
 author={Leviathan, Yaniv and Kalman, Matan and Matias, Yossi},
 booktitle={Proceedings of the 40th International Conference on Machine Learning},
 series={Proceedings of Machine Learning Research}, volume={202},
 pages={19274--19286}, year={2023}, publisher={PMLR},
 url={https://proceedings.mlr.press/v202/leviathan23a.html}
}

@inproceedings{chen2026dflash,
 title={{DFlash}: Block Diffusion for Flash Speculative Decoding},
 author={Chen, Jian and Liang, Yesheng and Liu, Zhijian},
 booktitle={Proceedings of the 43rd International Conference on Machine Learning},
 series={Proceedings of Machine Learning Research}, volume={306}, publisher={PMLR},
 year={2026}, url={https://arxiv.org/abs/2602.06036}
}

@misc{cheng2026dspark,
 title={{DSpark}: Confidence-Scheduled Speculative Decoding with Semi-Autoregressive Generation},
 author={Cheng, Xin and Yu, Xingkai and Shao, Chenze and Li, Jiashi and Xiong, Yunfan and Qian, Yi and Zhu, Jiaqi and Ma, Shirong and Zhang, Xiaokang and Ye, Jiasheng and Chen, Qinyu and Deng, Chengqi and Yu, Jiping and Dai, Damai and Zhang, Zhengyan and Wei, Yixuan and Tan, Yixuan and Yang, Wenkai and Xu, Runxin and Wu, Yu and Xu, Zhean and Wang, Xuanyu and Chen, Muyang and Tian, Rui and Bi, Xiao and Hao, Zhewen and Chen, Shaoyuan and Cao, Huanqi and Zhang, Wentao and Xu, Anyi and Zhang, Huishuai and Zhao, Dongyan and Liang, Wenfeng},
 year={2026}, eprint={2607.05147}, archivePrefix={arXiv}, howpublished={arXiv preprint arXiv:2607.05147},
 url={https://arxiv.org/abs/2607.05147}
}

@inproceedings{yang2025gated,
 title={Gated Delta Networks: Improving {Mamba2} with Delta Rule},
 author={Yang, Songlin and Kautz, Jan and Hatamizadeh, Ali},
 booktitle={International Conference on Learning Representations}, year={2025},
 url={https://openreview.net/forum?id=r8H7xhYPwz}
}

@misc{saxena2025cascade,
 title={Utility-Driven Speculative Decoding for Mixture-of-Experts},
 author={Saxena, Anish and Tsai, Po-An and Taneja, Hritvik and Jaleel, Aamer and Qureshi, Moinuddin},
 year={2025}, eprint={2506.20675}, archivePrefix={arXiv}, howpublished={arXiv preprint arXiv:2506.20675},
 url={https://arxiv.org/abs/2506.20675}
}

@misc{pan2026evict,
 title={Making Every Verified Token Count: Adaptive Verification for {MoE} Speculative Decoding},
 author={Pan, Lehan and Tao, Ziyang and Wang, Xiao and Zhao, Zhimeng and Wang, Meng and Zhang, Yanyong},
 year={2026}, eprint={2605.00342}, archivePrefix={arXiv}, howpublished={arXiv preprint arXiv:2605.00342},
 url={https://arxiv.org/abs/2605.00342}
}

@misc{xie2026ecospec,
 title={Less Experts, Faster Decoding: Cost-Aware Speculative Decoding for Mixture-of-Experts},
 author={Xie, Jincheng and Liu, Runheng and Huang, Heyan and Ling, Yawen and Dai, Hanbin and Zheng, Yu and Hu, Wen},
 year={2026}, eprint={2607.12696}, archivePrefix={arXiv}, howpublished={arXiv preprint arXiv:2607.12696},
 url={https://arxiv.org/abs/2607.12696}
}

@misc{wang2026specla,
 title={{SpecLA}: Efficient Speculative Decoding for Linear-Attention Models},
 author={Wang, Zhibin and Han, Xuying and Yang, Zhaohua and Liu, Fuliang and Li, Xue and Gu, Rong and Zhong, Sheng and Tian, Chen},
 year={2026}, eprint={2607.16673}, archivePrefix={arXiv}, howpublished={arXiv preprint arXiv:2607.16673},
 url={https://arxiv.org/abs/2607.16673}
}

@misc{wang2026bole,
 title={{Bole}: Efficient Tree Speculation for Hybrid-Attention Language Models},
 author={Wang, Li and Su, Yi and Wu, Xiabao and You, Chiran and Liu, Yongchao and Qiu, Zhan and Zhang, Juelu and Zheng, Jiajun and Liu, Fangxin and Zhang, Jie and Tian, Chen and Huan, Chengying},
 year={2026}, eprint={2608.01651}, archivePrefix={arXiv}, howpublished={arXiv preprint arXiv:2608.01651},
 url={https://arxiv.org/abs/2608.01651}
}

@misc{ghantasala2026treewy,
 title={{TreeWY}: Speculative Verification for {Gated DeltaNet} Hybrids},
 author={Ghantasala, Sneha Murthy},
 year={2026}, eprint={2608.20961}, archivePrefix={arXiv}, howpublished={arXiv preprint arXiv:2608.20961},
 url={https://arxiv.org/abs/2608.20961}
}

@inproceedings{katharopoulos2020linear,
 title={Transformers are {RNNs}: Fast Autoregressive Transformers with Linear Attention},
 author={Katharopoulos, Angelos and Vyas, Apoorv and Pappas, Nikolaos and Fleuret, Fran{\c{c}}ois},
 booktitle={Proceedings of the 37th International Conference on Machine Learning},
 series={Proceedings of Machine Learning Research}, volume={119},
 pages={5156--5165}, year={2020}, publisher={PMLR},
 url={https://proceedings.mlr.press/v119/katharopoulos20a.html}
}

@inproceedings{yang2024parallel,
 title={Parallelizing Linear Transformers with the Delta Rule over Sequence Length},
 author={Yang, Songlin and Wang, Bailin and Zhang, Yu and Shen, Yikang and Kim, Yoon},
 booktitle={Advances in Neural Information Processing Systems},
 volume={37}, year={2024},
 pages={115491--115522}, doi={10.52202/079017-3668}
}

@misc{chen2023sampling,
 title={Accelerating Large Language Model Decoding with Speculative Sampling},
 author={Chen, Charlie and Borgeaud, Sebastian and Irving, Geoffrey and Lespiau, Jean-Baptiste and Sifre, Laurent and Jumper, John},
 year={2023}, eprint={2302.01318}, archivePrefix={arXiv}, howpublished={arXiv preprint arXiv:2302.01318},
 url={https://arxiv.org/abs/2302.01318}
}

@inproceedings{li2025eagle3,
 title={{EAGLE-3}: Scaling up Inference Acceleration of Large Language Models via Training-Time Test},
 author={Li, Yuhui and Wei, Fangyun and Zhang, Chao and Zhang, Hongyang},
 booktitle={Advances in Neural Information Processing Systems},
 volume={38}, year={2025},
 pages={151568--151587}, doi={10.52202/085713-4562}
}

@inproceedings{dao2024ssm,
 title={Transformers are {SSM}s: Generalized Models and Efficient Algorithms Through Structured State Space Duality},
 author={Dao, Tri and Gu, Albert},
 booktitle={Proceedings of the 41st International Conference on Machine Learning},
 series={Proceedings of Machine Learning Research}, volume={235},
 pages={10041--10071}, year={2024}, publisher={PMLR},
 url={https://proceedings.mlr.press/v235/dao24a.html}
}

@inproceedings{wang2024mamballama,
 title={The {Mamba} in the {Llama}: Distilling and Accelerating Hybrid Models},
 author={Wang, Junxiong and Paliotta, Daniele and May, Avner and Rush, Alexander M. and Dao, Tri},
 booktitle={Advances in Neural Information Processing Systems},
 volume={37}, year={2024},
 pages={62432--62457}, doi={10.52202/079017-1996}
}

@inproceedings{wu2024snakes,
 title={Snakes and Ladders: Accelerating {SSM} Inference with Speculative Decoding},
 author={Wu, Yangchao and Dukler, Yonatan and Trager, Matthew and Achille, Alessandro and Xia, Wei and Soatto, Stefano},
 booktitle={Proceedings of The 4th NeurIPS Efficient Natural Language and Speech Processing Workshop},
 series={Proceedings of Machine Learning Research}, volume={262},
 pages={292--304}, year={2024}, publisher={PMLR},
 url={https://proceedings.mlr.press/v262/wu24a.html}
}

@inproceedings{wu2025stree,
 title={{STree}: Speculative Tree Decoding for Hybrid State Space Models},
 author={Wu, Yangchao and Qin, Zongyue and Wong, Alex and Soatto, Stefano},
 booktitle={Advances in Neural Information Processing Systems},
 volume={38}, pages={128121--128141}, year={2025}, doi={10.52202/085713-3855}
}

@misc{liou2026replayssm,
 title={{ReplaySSM}: Cache {SSM} Inputs, Not State},
 author={Liou, Ze-Wei and Dao, Tri},
 year={2026},
 howpublished={Dao AI Lab blog, June 15},
 url={https://dao-lab.ai/blog/2026/replayssm/}
}

@inproceedings{huang2025moesd,
 title={{MoESD}: Unveil Speculative Decoding's Potential for Accelerating Sparse {MoE}},
 author={Huang, Zongle and Zhu, Lei and Zhan, Zongyuan and Hu, Ting and Mao, Weikai and Yu, Xianzhi and Liu, Yongpan and Zhang, Tianyu},
 booktitle={Advances in Neural Information Processing Systems},
 volume={38}, year={2025},
 pages={138874--138909}, doi={10.52202/085713-4176}
}

@inproceedings{cai2024medusa,
 title={{Medusa}: Simple {LLM} Inference Acceleration Framework with Multiple Decoding Heads},
 author={Cai, Tianle and Li, Yuhong and Geng, Zhengyang and Peng, Hongwu and Lee, Jason D. and Chen, Deming and Dao, Tri},
 booktitle={Proceedings of the 41st International Conference on Machine Learning},
 series={Proceedings of Machine Learning Research}, volume={235},
 year={2024}, publisher={PMLR},
 url={https://proceedings.mlr.press/v235/cai24b.html}
}

@misc{hui2026peagle,
 title={{P-EAGLE}: Parallel-Drafting {EAGLE} with Scalable Training},
 author={Hui, Mude and Huang, Xin and Campos Salas, Jaime and Sun, Yue and Pemberton, Nathan and Song, Xiang and Khetan, Ashish and Karypis, George},
 year={2026}, eprint={2602.01469}, archivePrefix={arXiv}, howpublished={arXiv preprint arXiv:2602.01469},
 url={https://arxiv.org/abs/2602.01469}
}

@inproceedings{huang2024specdecpp,
 title={{SpecDec++}: Boosting Speculative Decoding via Adaptive Candidate Lengths},
 author={Huang, Kaixuan and Guo, Xudong and Wang, Mengdi},
 booktitle={Conference on Language Modeling ({COLM})},
 year={2025},
 url={https://openreview.net/forum?id=Y131N9fUbU}
}

@inproceedings{liu2025dropin,
 title={A Drop-In Solution for On-the-Fly Adaptation of Speculative Decoding in Large Language Models},
 author={Liu, Jiesong and Park, Brian and Shen, Xipeng},
 booktitle={Proceedings of the 63rd Annual Meeting of the Association for Computational Linguistics (Volume 1: Long Papers)},
 pages={9778--9794}, year={2025}, publisher={Association for Computational Linguistics},
 doi={10.18653/v1/2025.acl-long.482},
 url={https://aclanthology.org/2025.acl-long.482/}
}

@inproceedings{chen2024sequoia,
 title={{Sequoia}: Scalable and Robust Speculative Decoding},
 author={Chen, Zhuoming and May, Avner and Svirschevski, Ruslan and Huang, Yuhsun and Ryabinin, Max and Jia, Zhihao and Chen, Beidi},
 booktitle={Advances in Neural Information Processing Systems},
 volume={37}, year={2024},
 pages={129531--129563}, doi={10.52202/079017-4116}
}

@misc{chen2026faser,
 title={{FASER}: Fine-Grained Phase Management for Speculative Decoding in Dynamic {LLM} Serving},
 author={Chen, Wenyan and Lu, Chengzhi and Lin, Yanying and Ustiugov, Dmitrii},
 year={2026}, eprint={2604.20503}, archivePrefix={arXiv}, howpublished={arXiv preprint arXiv:2604.20503},
 url={https://arxiv.org/abs/2604.20503}
}

@misc{cobbe2021gsm8k,
 title={Training Verifiers to Solve Math Word Problems},
 author={Cobbe, Karl and Kosaraju, Vineet and Bavarian, Mohammad and Chen, Mark and Jun, Heewoo and Kaiser, Lukasz and Plappert, Matthias and Tworek, Jerry and Hilton, Jacob and Nakano, Reiichiro and Hesse, Christopher and Schulman, John},
 year={2021}, eprint={2110.14168}, archivePrefix={arXiv}, howpublished={arXiv preprint arXiv:2110.14168},
 url={https://arxiv.org/abs/2110.14168}
}

@inproceedings{zheng2024sglang,
 title={{SGLang}: Efficient Execution of Structured Language Model Programs},
 author={Zheng, Lianmin and Yin, Liangsheng and Xie, Zhiqiang and Sun, Chuyue and Huang, Jeff and Yu, Cody Hao and Cao, Shiyi and Kozyrakis, Christos and Stoica, Ion and Gonzalez, Joseph E. and Barrett, Clark and Sheng, Ying},
 booktitle={Advances in Neural Information Processing Systems},
 volume={37}, year={2024},
 pages={62557--62583}, doi={10.52202/079017-2000}
}

@misc{liu2026dcut,
 title={{D-cut}: Adaptive Verification Depth Pruning for Batched Speculative Decoding},
 author={Liu, Tianyu and Shen, Yuhao and Cen, Rui and Shi, Junhan and Zhang, Jiebin and Qin, Guangshuo and Liu, Hong and Liu, Song and Yu, Guanghua and Zhu, Jianchen},
 year={2026}, eprint={2607.14647}, archivePrefix={arXiv}, howpublished={arXiv preprint arXiv:2607.14647},
 url={https://arxiv.org/abs/2607.14647}
}

@inproceedings{abramovich2026speedbench,
 title={{SPEED-Bench}: A Unified and Diverse Benchmark for Speculative Decoding},
 author={Abramovich, Talor and Ashkenazi, Maor and Putterman, Izzy and Chislett, Benjamin and Mitra, Tiyasa and Darvish Rouhani, Bita and Zilberstein, Ran and Geifman, Yonatan},
 booktitle={Proceedings of the 43rd International Conference on Machine Learning},
 series={Proceedings of Machine Learning Research}, volume={306}, publisher={PMLR},
 year={2026}, url={https://arxiv.org/abs/2604.09557}
}

@misc{fu2026nemotronlabsdiffusion,
 title={{Nemotron-Labs-Diffusion}: A Tri-Mode Language Model Unifying Autoregressive, Diffusion, and Self-Speculation Decoding},
 author={Fu, Yonggan and Whalen, Lexington and Garg, Abhinav and Wu, Chengyue and Khadkevich, Maksim and Oswald, Nicolai and Xie, Enze and Egert, Daniel and Sreenivas, Sharath Turuvekere and Diao, Shizhe and Yu, Chenhan and Yu, Ye and Chen, Weijia and Norouzi, Sajad and Liu, Jingyu and Lan, Shiyi and Zhu, Ligeng and Wang, Jin and Jiang, Jindong and Mardani, Morteza and Maghoumi, Mehran and Han, Song and Juki{\'c}, Ante and Tajbakhsh, Nima and Kautz, Jan and Molchanov, Pavlo},
 year={2026}, eprint={2607.05722}, archivePrefix={arXiv}, howpublished={arXiv preprint arXiv:2607.05722},
 url={https://arxiv.org/abs/2607.05722}
}

@article{dinkelbach1967fractional,
 title={On Nonlinear Fractional Programming},
 author={Dinkelbach, Werner},
 journal={Management Science},
 volume={13}, number={7}, pages={492--498}, year={1967},
 url={https://doi.org/10.1287/mnsc.13.7.492}
}

@misc{qwen2026qwen35a3b,
 title={{Qwen3.5-35B-A3B}},
 author={{Qwen Team}},
 year={2026}, howpublished={Hugging Face model card},
 url={https://huggingface.co/Qwen/Qwen3.5-35B-A3B}
}

@misc{qwen2026qwen36a3b,
 title={{Qwen3.6-35B-A3B}},
 author={{Qwen Team}},
 year={2026}, howpublished={Hugging Face model card},
 url={https://huggingface.co/Qwen/Qwen3.6-35B-A3B}
}

@misc{qwen2026qwen35a10b,
 title={{Qwen3.5-122B-A10B-GPTQ-Int4}},
 author={{Qwen Team}},
 year={2026}, howpublished={Hugging Face model card},
 url={https://huggingface.co/Qwen/Qwen3.5-122B-A10B-GPTQ-Int4}
}

@inproceedings{tillet2019triton,
 title={Triton: An Intermediate Language and Compiler for Tiled Neural Network Computations},
 author={Tillet, Philippe and Kung, H. T. and Cox, David},
 booktitle={Proceedings of the 3rd ACM SIGPLAN International Workshop on Machine Learning and Programming Languages},
 pages={10--19}, year={2019}, publisher={ACM},
 doi={10.1145/3315508.3329973}
}

@inproceedings{schlag2021linear,
 title={Linear Transformers Are Secretly Fast Weight Programmers},
 author={Schlag, Imanol and Irie, Kazuki and Schmidhuber, J{\"u}rgen},
 booktitle={Proceedings of the 38th International Conference on Machine Learning},
 series={Proceedings of Machine Learning Research}, volume={139},
 pages={9355--9366}, year={2021}, publisher={PMLR},
 url={https://proceedings.mlr.press/v139/schlag21a.html}
}

@misc{lieber2024jamba,
 title={{Jamba}: A Hybrid {Transformer-Mamba} Language Model},
 author={Lieber, Opher and Lenz, Barak and Bata, Hofit and Cohen, Gal and Osin, Jhonathan and Dalmedigos, Itay and Safahi, Erez and Meirom, Shaked and Belinkov, Yonatan and Shalev-Shwartz, Shai and Abend, Omri and Alon, Raz and Asida, Tomer and Bergman, Amir and Glozman, Roman and Gokhman, Michael and Manevich, Avashalom and Ratner, Nir and Rozen, Noam and Shwartz, Erez and Zusman, Mor and Shoham, Yoav},
 year={2024}, eprint={2403.19887}, archivePrefix={arXiv}, howpublished={arXiv preprint arXiv:2403.19887},
 url={https://arxiv.org/abs/2403.19887}
}

@misc{minimax2025minimax01,
 title={{MiniMax-01}: Scaling Foundation Models with Lightning Attention},
 author={{MiniMax} and Li, Aonian and Gong, Bangwei and Yang, Bo and Shan, Boji and Liu, Chang and Zhu, Cheng and Zhang, Chunhao and Guo, Congchao and Chen, Da and Li, Dong and Jiao, Enwei and Li, Gengxin and Zhang, Guojun and Sun, Haohai and Dong, Houze and Zhu, Jiadai and Zhuang, Jiaqi and Song, Jiayuan and Zhu, Jin and Han, Jingtao and Li, Jingyang and Xie, Junbin and Xu, Junhao and Yan, Junjie and Zhang, Kaishun and Xiao, Kecheng and Kang, Kexi and Han, Le and Wang, Leyang and Yu, Lianfei and Feng, Liheng and Zheng, Lin and Chai, Linbo and Xing, Long and Ju, Meizhi and Chi, Mingyuan and Zhang, Mozhi and Huang, Peikai and Niu, Pengcheng and Li, Pengfei and Zhao, Pengyu and Yang, Qi and Xu, Qidi and Wang, Qiexiang and Wang, Qin and Li, Qiuhui and Leng, Ruitao and Shi, Shengmin and Yu, Shuqi and Li, Sichen and Zhu, Songquan and Huang, Tao and Liang, Tianrun and Sun, Weigao and Sun, Weixuan and Cheng, Weiyu and Li, Wenkai and Song, Xiangjun and Su, Xiao and Han, Xiaodong and Zhang, Xinjie and Hou, Xinzhu and Min, Xu and Zou, Xun and Shen, Xuyang and Gong, Yan and Zhu, Yingjie and Zhou, Yipeng and Zhong, Yiran and Hu, Yongyi and Fan, Yuanxiang and Yu, Yue and Yang, Yufeng and Li, Yuhao and Huang, Yunan and Li, Yunji and Huang, Yunpeng and Xu, Yunzhi and Mao, Yuxin and Li, Zehan and Li, Zekang and Tao, Zewei and Ying, Zewen and Cong, Zhaoyang and Qin, Zhen and Fan, Zhenhua and Yu, Zhihang and Jiang, Zhuo and Wu, Zijia},
 year={2025}, eprint={2501.08313}, archivePrefix={arXiv}, howpublished={arXiv preprint arXiv:2501.08313},
 url={https://arxiv.org/abs/2501.08313}
}

@misc{kimiteam2025kimilinear,
 title={{Kimi Linear}: An Expressive, Efficient Attention Architecture},
 author={{Kimi Team} and Zhang, Yu and Lin, Zongyu and Yao, Xingcheng and Hu, Jiaxi and Meng, Fanqing and Liu, Chengyin and Men, Xin and Yang, Songlin and Li, Zhiyuan and Li, Wentao and Lu, Enzhe and Liu, Weizhou and Chen, Yanru and Xu, Weixin and Yu, Longhui and Wang, Yejie and Fan, Yu and Zhong, Longguang and Yuan, Enming and Zhang, Dehao and Zhang, Yizhi and Liu, T. Y. and Wang, Haiming and Fang, Shengjun and He, Weiran and Liu, Shaowei and Li, Yiwei and Su, Jianlin and Qiu, Jiezhong and Pang, Bo and Yan, Junjie and Jiang, Zhejun and Huang, Weixiao and Yin, Bohong and You, Jiacheng and Wei, Chu and Wang, Zhengtao and Hong, Chao and Chen, Yutian and Chen, Guanduo and Wang, Yucheng and Zheng, Huabin and Wang, Feng and Liu, Yibo and Dong, Mengnan and Zhang, Zheng and Pan, Siyuan and Wu, Wenhao and Wu, Yuhao and Guan, Longyu and Tao, Jiawen and Fu, Guohong and Xu, Xinran and Wang, Yuzhi and Lai, Guokun and Wu, Yuxin and Zhou, Xinyu and Yang, Zhilin and Du, Yulun},
 year={2025}, eprint={2510.26692}, archivePrefix={arXiv}, howpublished={arXiv preprint arXiv:2510.26692},
 url={https://arxiv.org/abs/2510.26692}
}

@inproceedings{miao2024specinfer,
 title={{SpecInfer}: Accelerating Large Language Model Serving with Tree-based Speculative Inference and Verification},
 author={Miao, Xupeng and Oliaro, Gabriele and Zhang, Zhihao and Cheng, Xinhao and Wang, Zeyu and Zhang, Zhengxin and Wong, Rae Ying Yee and Zhu, Alan and Yang, Lijie and Shi, Xiaoxiang and Shi, Chunan and Chen, Zhuoming and Arfeen, Daiyaan and Abhyankar, Reyna and Jia, Zhihao},
 booktitle={Proceedings of the 29th ACM International Conference on Architectural Support for Programming Languages and Operating Systems, Volume 3},
 pages={932--949}, year={2024}, publisher={ACM},
 doi={10.1145/3620666.3651335}
}

@inproceedings{li2024eagle2,
 title={{EAGLE-2}: Faster Inference of Language Models with Dynamic Draft Trees},
 author={Li, Yuhui and Wei, Fangyun and Zhang, Chao and Zhang, Hongyang},
 booktitle={Proceedings of the 2024 Conference on Empirical Methods in Natural Language Processing},
 pages={7421--7432}, year={2024}, publisher={Association for Computational Linguistics},
 doi={10.18653/v1/2024.emnlp-main.422},
 url={https://aclanthology.org/2024.emnlp-main.422/}
}

@inproceedings{kwon2023vllm,
 title={Efficient Memory Management for Large Language Model Serving with {PagedAttention}},
 author={Kwon, Woosuk and Li, Zhuohan and Zhuang, Siyuan and Sheng, Ying and Zheng, Lianmin and Yu, Cody Hao and Gonzalez, Joseph and Zhang, Hao and Stoica, Ion},
 booktitle={Proceedings of the 29th Symposium on Operating Systems Principles},
 pages={611--626}, year={2023}, publisher={ACM},
 doi={10.1145/3600006.3613165}
}

@misc{nvidia2026cudagraphs,
 title={{CUDA} Programming Guide, Section 4.2: {CUDA Graphs}},
 author={{NVIDIA}},
 year={2026}, howpublished={Online documentation, last updated September 10},
 url={https://docs.nvidia.com/cuda/cuda-programming-guide/04-special-topics/cuda-graphs.html}
}
